%% file: main.tex
\documentclass{article}

\usepackage{microtype}
\usepackage{graphicx}
\usepackage{subcaption}
\usepackage{booktabs} % for professional tables

\usepackage{hyperref}

\usepackage[table, dvipsnames]{xcolor}
\usepackage{bm}
\usepackage[linesnumbered,ruled]{algorithm2e}
\expandafter\def\csname ver@algorithm.sty\endcsname{}
\usepackage{algorithmic}
\usepackage{hyperref}
\usepackage[capitalise]{cleveref}
\crefname{section}{§}{§§}
\Crefname{section}{§}{§§}
\crefformat{section}{\S#2#1#3}
\crefname{figure}{Fig}{Figs}
\Crefname{figure}{Fig}{Figs}
\crefname{subfigure}{Fig}{Figs}
\Crefname{subfigure}{Fig}{Figs}
\crefname{problem}{Prob}{Probs}
\Crefname{problem}{Prob}{Probs}
\usepackage{subfiles}
\usepackage{pgfplots} % figures/fig_*.tex
\pgfplotsset{compat=1.16}
\usetikzlibrary{calc,fit,decorations.pathreplacing} % figures/fig_sharing.tex
\usepackage{xspace}
\newcommand{\sys}{Readie\xspace}

\usepackage[accepted]{mlsys2025}
\usepackage[varqu,varl]{inconsolata} % \texttt

\makeatletter
\def\mlsys@tabletype{table}
\long\def\@makecaption#1#2{%
  \ifx\@captype\mlsys@tabletype\else\vskip 10pt\fi
  \baselineskip 11pt
  \setbox\@tempboxa\hbox{#1. #2}%
  \ifdim \wd\@tempboxa >\hsize
    \sbox{\newcaptionbox}{\small\sl #1.~}%
    \newcaptionboxwid=\wd\newcaptionbox
    \usebox\newcaptionbox {\footnotesize #2}%
  \else
    \centerline{{\small\sl #1.} {\small #2}}%
  \fi
  \ifx\@captype\mlsys@tabletype\vskip 1mm\fi}
\makeatother

\makeatletter
\def\paragraph{\@startsection{paragraph}{4}{\z@}{0.3ex plus 0.2ex minus .1ex}{-1em}{\normalsize\bfseries\itshape}}
\def\@listi{\leftmargin\leftmargini
  \topsep 2pt plus 1pt minus 1pt \partopsep 0pt
  \itemsep 1pt plus 0.5pt minus 0.5pt \parsep 0pt}
\let\@listI\@listi
\@listi
\g@addto@macro\normalsize{%
  \setlength\abovedisplayskip{5pt plus 2pt minus 2pt}%
  \setlength\belowdisplayskip{5pt plus 2pt minus 2pt}%
  \setlength\abovedisplayshortskip{2pt plus 1pt}%
  \setlength\belowdisplayshortskip{3pt plus 1pt minus 1pt}}
\let\mlsysabstract\abstract
\def\abstract{\setlength{\parskip}{6pt}%
  \def\@listi{\leftmargin\leftmargini \topsep 4pt plus 1pt minus 2pt \partopsep 1pt plus 0.5pt minus 0.5pt}%
  \mlsysabstract}
\makeatother
\usepackage{enumitem}
\setlist[itemize]{leftmargin=*, topsep=0pt, partopsep=0pt}

\newcommand{\mytitle}{\sys: Low-Latency Code Sandboxes for AI Agents}

\mlsystitlerunning{\mytitle}

\begin{document}

\twocolumn[
\mlsystitle{\mytitle}

\subfile{header/authors}

\subfile{header/abstract}
]



\printAffiliationsAndNotice{} % otherwise use the standard text.

%%% sections %%%
\subfile{sections/introduction}
\subfile{sections/background}
\subfile{sections/system_overview}
\subfile{sections/formulation}
\subfile{sections/implementation}
\subfile{sections/experiments}
\subfile{sections/related_work}

\subfile{sections/conclusion}
%%% end sections %%%

\section*{Acknowledgements}

This project is supported in part by the NSF grants \#2312561 and \#~2440498, and by resources from the Advanced Cyberinfrastructure Coordination Ecosystem: Services \& Support (ACCESS) program, which is supported by NSF grants \#2138259, \#2138286, \#2138307, \#2137603, and \#2138296.
Eugene Wu is partially supported by DAPLab corporate support in the form of funding and/or compute from Amazon, IntellectAI, Infosys, Tidalwave, Veris, shopify, Microsoft, Thinking Machines, Dandy, Perplexity, and Daytona.
The views and conclusions presented here are those of the authors and should not be interpreted as representing the official positions of the funding organizations.


\bibliography{main}
\bibliographystyle{mlsys2025}

%%%%%%%%%%%%%%%%%%%%%%%%%%%%%%%%%%%%%%%%%%%%%%%%%%%%%%%%%%%%%%%%%%%%%%%%%%%%%%%
%%%%%%%%%%%%%%%%%%%%%%%%%%%%%%%%%%%%%%%%%%%%%%%%%%%%%%%%%%%%%%%%%%%%%%%%%%%%%%%
% SUPPLEMENTAL CONTENT AS APPENDIX AFTER REFERENCES
%%%%%%%%%%%%%%%%%%%%%%%%%%%%%%%%%%%%%%%%%%%%%%%%%%%%%%%%%%%%%%%%%%%%%%%%%%%%%%%
%%%%%%%%%%%%%%%%%%%%%%%%%%%%%%%%%%%%%%%%%%%%%%%%%%%%%%%%%%%%%%%%%%%%%%%%%%%%%%%
% \appendix
% \section{Please add supplemental material as appendix here}
% %
% Put anything that you might normally include after the references as an appendix here, {\it not in a separate supplementary file}. Upload your final camera-ready as a single pdf, including all appendices.

%%%%%%%%%%%%%%%%%%%%%%%%%%%%%%%%%%%%%%%%%%%%%%%%%%%%%%%%%%%%%%%%%%%%%%%%%%%%%%%
%%%%%%%%%%%%%%%%%%%%%%%%%%%%%%%%%%%%%%%%%%%%%%%%%%%%%%%%%%%%%%%%%%%%%%%%%%%%%%%

\end{document}

%% file: header/authors.tex
% unless you've provided the [accepted] option to the mlsys2025


% \mlsyssetsymbol{equal}{*}

\begin{mlsysauthorlist}
\mlsysauthor{Arjun Sivaraman}{illinois}
\mlsysauthor{Zhaoheng Li}{bytedance}
\mlsysauthor{Alan Shi}{illinois}
\mlsysauthor{Sanjay Malisetty}{illinois}
\mlsysauthor{Eugene Wu}{columbia}
\mlsysauthor{Yongjoo Park}{illinois}
\end{mlsysauthorlist}

\mlsysaffiliation{illinois}{University of Illinois Urbana-Champaign}
\mlsysaffiliation{bytedance}{ByteDance}
\mlsysaffiliation{columbia}{Columbia University}

\mlsyscorrespondingauthor{Arjun Sivaraman}{arjun16@illinois.edu}
\mlsyscorrespondingauthor{Yongjoo Park}{yongjoo@illinois.edu}

%% file: header/abstract.tex
% \mlsyskeywords{Code Sandboxes, AI Agents, Cold Start, Serverless Computing, Memory Snapshots, Checkpoint-Restore}

\vskip 0.3in

\begin{abstract}

From quantitative research to personal assistance, AI agents act by iteratively generating and executing code.
To decouple the agent harness from code execution, many agents offload execution to dedicated systems called \emph{code sandboxes}.
However, existing code sandboxes suffer from slow startup.
Whether they rely on custom snapshots or a single fat image, the root cause is the same: it is hard to build a compact, pre-warmed snapshot that can be shared across diverse agent-generated code while covering nearly all the artifacts (e.g., packages, datasets) it requires.
In this work, we propose \emph{learned startup snapshots} for low-latency code sandboxes.
Our insight is that while it is hard for a single snapshot to be both compact and cover most workloads, a set of snapshots can achieve both.
Each snapshot is small, initializes quickly, and covers a fraction of agent-generated code; together, the set covers most of it.
We learn this set from a large training workload by formulating snapshot construction as an optimization problem and solving it with an approximation algorithm.
Planned on 15K public and LLM-synthesized scripts, our system, \sys, reduces startup latency by 3--22$\times$ and end-to-end runtime by 1.3--3.3$\times$ compared to existing approaches at the same storage budget.

\end{abstract}

%% file: sections/introduction.tex
\section{Introduction}
\label{sec:intro}

AI agents rely heavily on code execution. To give financial advice, an agent may call \citet{duckduckgo} to retrieve current mortgage rates and US Treasury yields. To perform quantitative political science research, it may use \citet{pandas} over government datasets. And to shrink vacation photos before emailing them to family, it may use \citet{pillow} to compress hundreds of images.
To run such code at scale, independent of the agent's development environment, many agents offload execution to dedicated systems called \emph{code sandboxes}~\cite{modal,e2b,daytona}. Sandbox latency directly affects agents' task completion time: the faster, the better.

Existing sandboxes incur long startup times to provision required artifacts such as libraries/packages, datasets, etc.
For example, to run a pandas script, the sandbox must include the pandas package.
Existing approaches address this through: (i) custom container images (\cref{fig:intro}-bottom left), (ii) custom snapshots (\cref{fig:intro}-bottom right), or (iii) a shared fat image (\cref{fig:intro}-top left).
First, services such as Cloud Run~\cite{cloudrun}\footnote{%
    We use Cloud Run, Modal, and Colab as examples based on their current offerings, without considering their business positioning. 
    They may develop similar techniques in the future.
} let users build custom container images, and others such as Modal~\cite{modal_memory_snapshots} let users build snapshots containing useful artifacts.
However, these images or snapshots must be fetched from remote storage unless they are cached from a previous run.
Moreover, hand-built images or snapshots may miss some artifacts, incurring slow runtime downloads (e.g., \texttt{pip install}).
Second, Colab-like~\cite{colab-backend-info} platforms use a single fat image of commonly used packages (e.g., \citet{numpy}, \citet{scikit-learn}).
This image is static: its contents are chosen in advance rather than adapted to usage.
Furthermore, loading artifacts from a file-based image is relatively slower than memory-state snapshots.

\input{figures/fig_intro.tex}

In this work, we propose \textbf{\emph{learned startup snapshots}} for low-latency code sandboxes.
The core idea is that while it is hard for a single snapshot to include all required artifacts, it is far easier to ensure that at least one of several snapshots does.
Because each snapshot covers only a subset of artifacts, it stays small (a few tens of megabytes) and initializes quickly.
Our system, \sys, serves requests as follows:
Given a script, \sys finds the smallest snapshot that contains most required artifacts by evaluating its internal cost function on expected startup time; 
then, \sys clones an isolated container from the snapshot.
It then executes the script and destroys the container.
Unlike custom snapshots, \sys learns which artifacts to include from usage, so users need not predict them.
Unlike a shared fat image, \sys skips loading artifacts at startup, since they are already in the snapshot's memory.

In designing \sys, a core research problem is to build a set of snapshots that minimizes the expected startup cost.
One naive approach is to build a single snapshot from the most popular artifacts (e.g., based on statistics from public scripts).
However, to cover most scripts, this snapshot must include many artifacts, which makes it too large to load quickly.
Another naive approach is to mine artifacts that are frequently used together~\cite{agrawal1994fast, han2000mining,HotSwap}.
However, we find that this frequency-based approach ignores the actual cost of loading each artifact.
Instead, we formulate an optimization problem whose objective models the actual load time of each artifact, considering the dependencies among artifacts.
Its solution is a near-optimal set of snapshots within a storage budget:
the formulation is exact, but we solve it with an approximation algorithm for scalability.

\sys targets (i) fast execution of previously unseen code with (ii) container-level isolation, achieving both efficiency and security.
This differs from existing work.
Per-function snapshots~\cite{Prebaking,Pronghorn,Shin2022Fireworks} are created in advance for each function, so they are not optimized for new code.
State-sharing systems~\cite{SOCK,chai2025fork,yang2024forklift,li2021pagurus,HotSwap} decide heuristically what to share.
They also let different requests share live in-memory state, which can weaken isolation between them.
Other cold-start optimization techniques are orthogonal to ours (\cref{sec:related}).

This work makes the following contributions:
\begin{enumerate}[topsep=0pt]
\item We show why existing approaches are poorly suited for agent workloads, and why learned startup snapshots are effective (\cref{sec:motivation}).
\item We present the design of \sys, a practical system that serves requests with learned startup snapshots (\cref{sec:system}).
\item We formulate snapshot construction as an optimization problem and present an approximation algorithm that finds a near-optimal set of snapshots (\cref{sec:formulation}).
\item We describe the implementation challenges we addressed in building \sys (\cref{sec:implementation}).
\item We evaluate our system (\sys) on 15K scripts: it reduces both startup latency and end-to-end time (\cref{sec:evaluation}).
\end{enumerate}

Finally, we discuss related work (\cref{sec:related}) and conclude (\cref{sec:conclusion}).

%% file: figures/fig_intro.tex
% Quadrant of startup approaches: file vs. memory (x), customized vs. shareable (y).
% Origin at the bottom left. Black and white only.
% Use: \input{figures/fig_intro.tex}; needs tikz (loaded by pgfplots).
\begin{figure}[t]
\centering
\begin{tikzpicture}[font=\small]
  \def\w{5}
  \def\h{4}
  \draw[thin] (\w/2,0) -- (\w/2,\h);
  \draw[thin] (0,\h/2) -- (\w,\h/2);
  \draw[thick] (0,0) rectangle (\w,\h);
  \node[anchor=north] at (\w/4,0)   {Files};
  \node[anchor=north] at (3*\w/4,0) {Memory-state};
  \node[anchor=south, rotate=90] at (-0.1,\h/4)   {Customized};
  \node[anchor=south, rotate=90] at (-0.1,3*\h/4) {Shareable};
  \node[align=center] at (\w/4,3*\h/4) {A fat image\\(e.g., Colab)};
  \node[align=center] at (\w/4,\h/4) {Custom container\\images\\(e.g., Cloud Run)};
  \node[align=center] at (3*\w/4,\h/4) {Hand-crafted\\Snapshots\\ (e.g., Modal)};
  \node[align=center] at (3*\w/4,3*\h/4) {Learned\\Snapshots\\\textbf{(Ours)}};
\end{tikzpicture}

\vspace{-3mm}
\caption{We leverage \emph{learned startup snapshots} to achieve low-latency code sandboxes.
That is, we create a set of snapshots: each is small in size (thus clones fast) 
  while at least one of them likely contains required artifacts, 
  thus offering high coverage.}
\label{fig:intro}
\end{figure}

%% file: sections/background.tex
\section{Motivation}
\label{sec:motivation}

This section discusses why existing sandboxes start slowly (\cref{sec:motivation_current_methods}), our intuition for using only a few snapshots to serve a large number of requests (\cref{sec:motivation_snapshot}), and the characteristics of agent-generated workloads that enable it (\cref{sec:motivation_characteristics}).

\begin{figure}[t]
    \centering
    \input{figures/fig_agents_trajectories.tex}
    \vspace{-2mm}
    \caption{New packages that each of 10 agent sessions requests for the first time at each turn (light blue), and their mean at each turn (red), rounded up. It also shows the actual packages requested in one of the trajectories (solid blue).}
    \label{fig:new_packages_per_turn}
\end{figure}
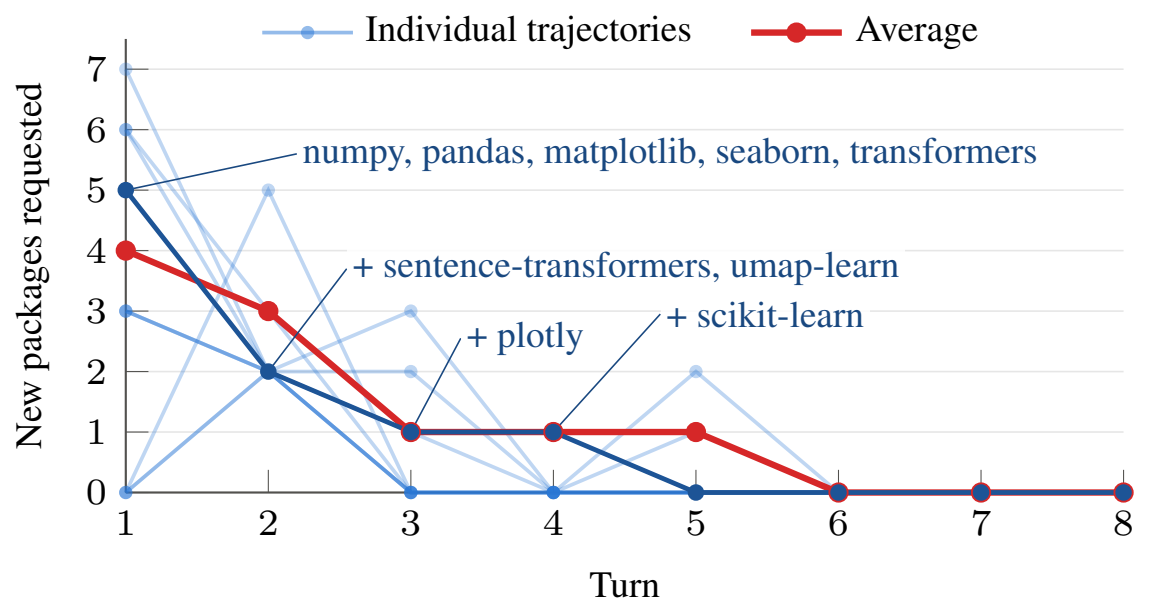

\subsection{Why Are Current Methods Slow?}
\label{sec:motivation_current_methods}

Pre-built images or snapshots can be fast if they include the required artifacts.
In practice, however, they often do not, and existing solutions fail to close this gap.
In this section, we support these claims with three empirical studies.
First, by analyzing agent-generated trajectories, we show that snapshots from previous runs are often not reusable for future runs.
Second, we show that downloading artifacts at request time, with only a base image cached on each VM, results in high startup latency.
Third, we show that the existing image-sharing approach, a single fat image, also incurs slow startup.

% Sandboxes prepare the environment that the code needs in one of the following ways: they build a custom image/snapshot for each run, or they share one fat image across all runs. Since each run can need a different set of packages, a custom image built for a previous run may not be optimal for the next one, and hence a new image is needed for every new set of requirements. Loading such an image into a sandbox takes significant time, because it is unique to its requirements, and so it is not frequently cached. A fat image avoids this by installing the most commonly used packages in a single image that every run shares, so the packages a run needs are usually already on disk. However, every new sandbox must still import its packages from disk, which takes time that grows with how much code each package loads and runs.

\begin{figure}[t]
    \centering
    \input{figures/fig_layer_pull.tex}\\[10pt]
    \input{figures/fig_import_phases.tex}
    \caption{Analysis of startup cost.
    % Time until pandas, PyTorch and NLTK are ready to use: 
    (a) Loading a pre-built image becomes slower if layers need to be downloaded or artifacts are missing;
    % (a) in a custom image, with every layer pulled from a registry over a 1\,Gbit/s link, 
    (b) If artifacts reside on disks, cloning/restoring from a memory-state snapshot is faster than reading artifacts (i.e., Python packages here) from original source code or cached bytecode.
    % in a gVisor sandbox, importing each package from disk, from pre-compiled bytecode, or from a restored snapshot.
    }
    \label{fig:layer_pull}
\end{figure}

\paragraph{Packages the code needs are not known in advance.}
We study whether the packages an agent needs for an entire session are known from the code executed in its first turn, by recording the packages imported at each turn. We use 6,833 agent sessions, in which an agent runs code at each turn to work on a task: trajectories recorded on agent benchmarks, general data science tasks, and code debugging sessions (see \cref{sec:evaluation_experiment_setup} and \cref{sec:evaluation_agent} for details). In 30.8\% of these sessions, agents import new packages after their first turn, and they keep requesting new packages until turn 3.7 on average (\cref{fig:new_packages_per_turn}).
An agent requests new packages after its first turn because it decides each step only after seeing the previous step’s output~\cite{hetero_batching_retrieval}. The session drawn in solid-blue in \cref{fig:new_packages_per_turn} first classifies the sentiment of product reviews with \citet{transformers}, pandas, and \citet{seaborn}. It then embeds the reviews (\citet{sentence-transformers}, \citet{umap-learn}), plots them (\citet{plotly}), and finally models their topics (scikit-learn).

\paragraph{Using a custom image for each execution.}
Services such as Cloud Run~\cite{cloudrun} and AWS Lambda~\cite{lambda-container-loading} initialize each sandbox from a container image that holds the code’s packages.  We measure how long such a sandbox takes to become ready to execute code when its image is not cached on the host.
We pulled an image with a base layer, a runtime layer, and a custom layer that installs pandas, \citet{pytorch} and \citet{nltk} from a remote registry over a 1\,Gbit/s link, with no layer cached. We then started a container from the downloaded image and ran code that imports the three packages (\cref{fig:layer_pull_a}). The sandbox was ready to use after 17.4\,s, of which downloading and extracting the custom layer took 15.0\,s, and starting the container and importing the packages took 2.1\,s.

\paragraph{Using one fat image for all executions.}
Notebook services such as Colab~\cite{colab-backend-info} reduce request-time downloads by running every sandbox from a large shared image with common packages installed. 
We measure how long a fresh sandbox takes to import packages that the code requires from such an image, and how much of that time a snapshot saves.
On the ~\citet{kaggle-docker-python} image, which contains many commonly used packages, 
we imported each of pandas, PyTorch and NLTK separately in three ways: from disk with no compiled bytecode, from bytecode compiled ahead of time, and by restoring a snapshot taken after the import (\cref{fig:layer_pull_b}).
From disk, the imports took 2.74--7.28\,s, of which 56--70\% was bytecode compilation time. With compiled bytecode, importing still took 1.00--2.80\,s to load and run the modules. Restoring the snapshot skipped both steps, and the sandbox was ready in 0.29--0.42\,s.

\subsection{Missed Opportunity: Learned Snapshots}
\label{sec:motivation_snapshot}

We hypothesize that a \emph{small} number of \emph{compact} snapshots can \emph{cover} a large number of requests (i.e., scripts to execute).
This hypothesis rests on three elements: a definition of \emph{coverage}, an effective approach to \emph{compactness}, and a one-to-many mapping from snapshots to requests, which keeps the number of snapshots \emph{small}.
This subsection defines coverage, and then argues that compactness requires novel techniques.
The next subsection examines whether a one-to-many mapping holds in practice.

% \cref{sec:motivation_current_methods} shows that restoring a snapshot can significantly reduce the time required to set up a new sandbox compared to importing packages from source. Some platforms today already snapshot a process after startup and restore it at every execution request on a per-function or per-image basis.

\paragraph{Cost-based coverage.}
In principle, even an empty snapshot can cover any request by downloading required artifacts at request time.
To reflect this, we define coverage through a cost function.
Suppose a snapshot $S_i$ contains artifacts $a_1$, $a_2$, and $a_3$, and a request $R_j$ requires $a_1$ and $a_4$.
Then, serving $R_j$ with $S_i$ incurs the cost of loading $a_1$ from the snapshot (fast) and downloading $a_4$ (slow).
We call the missing artifacts, $R_j \setminus S_i = \{a_4\}$, the \emph{residuals}.

\begin{figure}[t]
    \centering
    \input{figures/fig_sharing.tex}
    \caption{Conceptual difference between (a) per-request snapshot vs (b) a shareable snapshot for multiple requests.
    We take the second approach through learned startup snapshots.}
    \label{fig:sharing}
\end{figure}

% A request can directly use any snapshot whose set of packages is a superset of the packages it needs. The packages it does not require sit unused in memory, but the benefits of using the snapshot are still realized. If a snapshot contains only a subset of packages that a request needs, it can still be used to serve the request as long as the remaining packages are imported after the snapshot is restored. We call the packages that are imported post-restore \emph{residual imports}.

\Cref{fig:sharing} contrasts per-request snapshots (\cref{fig:sharing_a}) with a single snapshot shared across requests (\cref{fig:sharing_b}).
With per-request snapshots, serving five requests, each requiring a different set of artifacts, requires five snapshots.
With our relaxed definition of coverage, a single snapshot can serve all five, though at different startup costs~\cite{lazyattention,cache_of_thought}.
For example, a 32\,MB snapshot containing NumPy, pandas, scikit-learn and \citet{matplotlib} fully covers $R_1$--$R_3$, which incur no additional load time.
Requests $R_4$ and $R_5$ must also load their residuals, seaborn and \citet{xgboost}, which takes 0.3\,s and 0.7\,s, respectively.
Even so, they still skip importing the four shared artifacts, which account for most of their import time.
Overall, one snapshot replaces five, cutting storage by 5$\times$ at the cost of only 1\,s of additional load time in total.

\paragraph{Compact snapshot through learning.} 

Finding an optimal set of snapshots, one that minimizes the total startup cost across all requests, requires more than simple heuristics.
In \cref{fig:sharing_b}, for example, including NumPy and pandas in a snapshot is an easy choice, since $R_1$--$R_5$ all require them.
Whether to also include scikit-learn, however, depends on the workload.
If $R_1$-like requests, which do not need scikit-learn, dominate the workload, including it would increase the snapshot load time for most requests, while saving residual load time for only a few.

To address it, we take a learning-based approach: we find optimal snapshots using a training workload~\cite{database_learning,keebo}.
While the training workload could be tailored to a specific deployment, we instead collect a large number of scripts from public sources (e.g., \citet{kaggle}, \citet{huggingface}) and synthesize more with frontier large language models~\cite{notebook_checkpointing_agents}.
Just like a pre-trained model generalizes to a wide range of tasks, 
we expect our snapshots to efficiently cover most practical workloads. 
We validate this in \cref{sec:evaluation}.

% The restore time of a snapshot grows with its size, i.e., the size of the packages it contains. A larger snapshot might reduce residual imports, but costs more time to restore. A smaller snapshot is faster to restore but requires more residual imports. A good set of shared snapshots balances the two.

\subsection{Characteristics of Workloads}
\label{sec:motivation_characteristics}

\begin{figure}[t]
    \centering
    \input{figures/fig_workload_clusters.tex}

    \vspace{-2mm}
    \caption{%
    Snapshots and requests have a one-to-many mapping: a few snapshots can cover many requests.
    A small dot is a code execution request, and a large dot is a candidate snapshot.
    An oval around a snapshot encloses the requests it can efficiently cover.
    % 382 distinct package sets (used by more than 3 requests) sized by their request frequency among 9,325 requests. 
    % The sets are clustered by Jaccard distance, and laid out with t-SNE.
    }
    \label{fig:workload_patterns}
\end{figure}
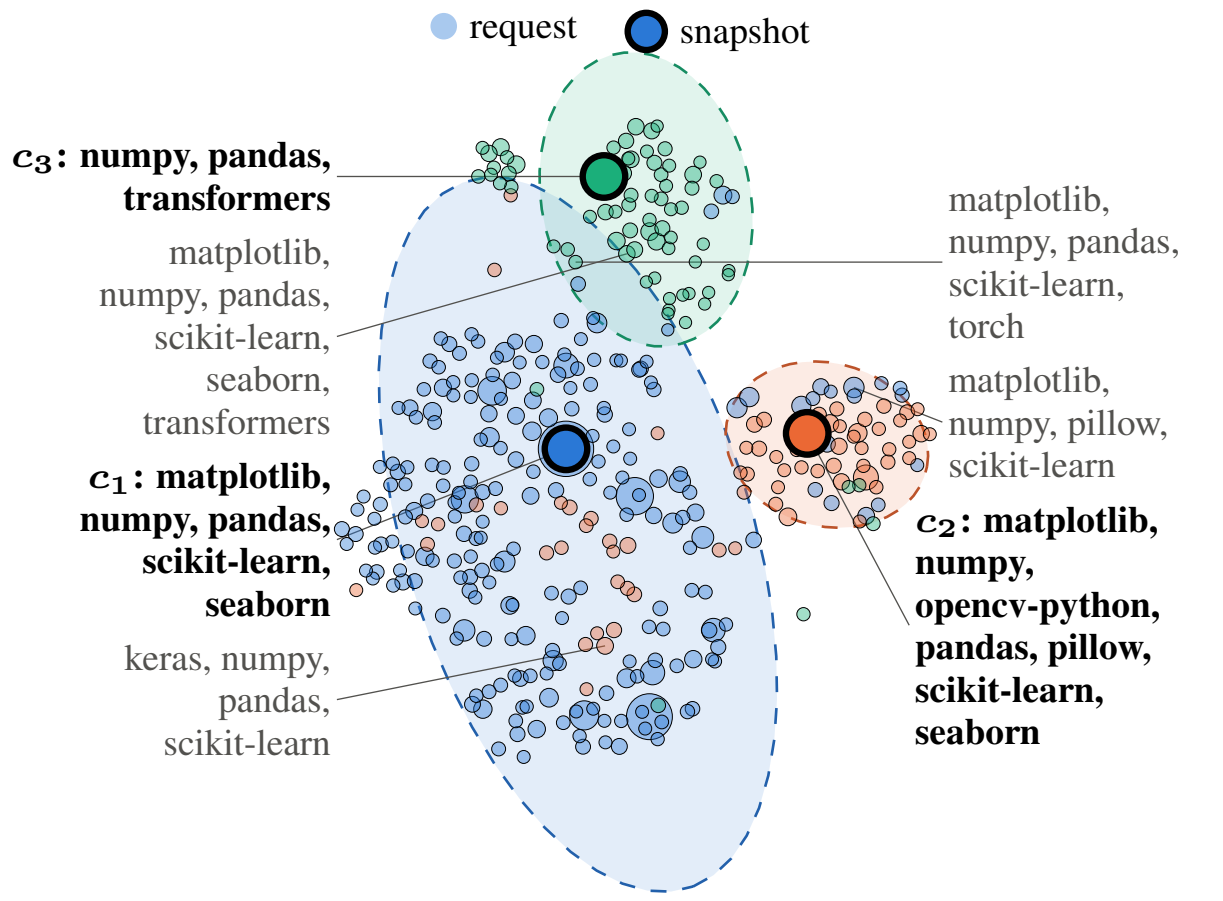

We examine whether one-to-many relationships hold between snapshots and requests. 
If so, only a few snapshots can cover a large number of requests efficiently.
We represent each of the 15,299 requests in our planning set (\cref{sec:evaluation_experiment_setup}) by the set of packages it imports, cluster these sets by Jaccard distance \cite{jaccard1901etude} with average linkage. The clusters hold 11,664 requests with 2,472 distinct package sets, and 75\% of which are required by only one request (\cref{fig:workload_patterns}). We pick one request near each cluster's center as a snapshot and price every request on it with the cost model defined in \cref{sec:formulation_initialization_time}.

\paragraph{One snapshot covers multiple requests.}
Each cluster shares a core set of packages: NumPy, pandas and scikit-learn (blue, 82\% of the requests); Pillow, \citet{opencv} and torchvision~\cite{pytorch} (orange); and Transformers, \citet{hf-datasets} and PyTorch (green). 
Of the 9,325 requests in \cref{fig:workload_patterns}, $c_1$ serves 83\%, and $c_2$ and $c_3$ serve 9\% and 7\%. Only 1,731 of $c_1$'s 7,775 requests match it exactly: 2,286 need a subset of its packages, and 1,500 import a few more as residuals. Snapshots that match requests exactly would instead need 1,306 snapshots and 351\,GB to serve 90\% of these requests. The most common set is not always the best center, however. The orange cluster's most common set, \{Pillow, OpenCV, Matplotlib, NumPy\}, serves 288 requests, while adding pandas and scikit-learn to it, as $c_2$ does, serves 853.

\paragraph{Overlaps create trade-offs and redundancy.} 
The request \{Matplotlib, NumPy, pandas, scikit-learn, PyTorch\} lies where $c_1$ and $c_3$ overlap. $c_1$ holds four of its five packages, but restoring it (0.44\,s) requires importing PyTorch from disk, which costs 14.8\,s. $c_3$ holds PyTorch, which Transformers imports, so restoring it takes 0.86\,s and serves the rest in 13.6\,s, in a lower time than $c_1$. Overlaps also make snapshots redundant: $c_2$ holds all of $c_1$ plus Pillow and OpenCV, and restores only 0.02\,s slower, so for 7,954 of the 9,325 requests the two are within 10\% of each other. Choosing snapshots therefore means weighing such overlaps against a storage budget, with a cost model rather than the clusters' shape alone (\cref{sec:formulation}).

%% file: figures/fig_agents_trajectories.tex
% New third-party packages that each of ten agent sessions requests at each turn (faint lines), and
% their mean at each turn, rounded up to a whole package (heavy line); a session that has ended counts as zero. One column wide; same axis
% style as fig_planners.tex. One session (cpu:agent_notebook[1432]) is drawn solid and labelled with
% the packages it requests at each turn.
% Use: \input{figures/fig_agents_trajectories.tex} inside a figure; needs \usepackage{pgfplots}.
% Data: figures/agents_trajectories.dat, written by experiments/new_packages_per_turn/export_trajectories.py.
\begingroup
\definecolor{atLine}{HTML}{2A78D6}%
\definecolor{atMean}{HTML}{D62728}%
\definecolor{atInk}{HTML}{52514E}%
\begin{tikzpicture}
  \begin{axis}[
      width=6.6cm, height=3.0cm, scale only axis,
      xmin=1, xmax=8, ymin=0, ymax=7.5,
      xtick={1,...,8}, ytick={0,1,...,7},
      ticklabel style={font=\scriptsize}, label style={font=\scriptsize},
      xlabel={Turn}, ylabel={New packages requested},
      axis lines*=left, axis line style={atInk}, tick style={atInk},
      ymajorgrids, grid style={black!10, line width=0.3pt},
      unbounded coords=jump,
      legend style={font=\scriptsize, draw=none, fill=none, at={(0.5,1.1)}, anchor=north, legend columns=2,
        /tikz/every even column/.append style={column sep=0.3cm}},
      legend cell align=left, clip=false,
    ]
    \foreach \s in {0,...,9} {
      \addplot[atLine, opacity=0.3, line width=0.7pt, mark=*, mark size=0.9pt, forget plot]
        table[x=turn, y=s\s] {figures/agents_trajectories.dat};
    }
    \addlegendimage{atLine, opacity=0.5, line width=0.7pt, mark=*, mark size=0.9pt}
    \addlegendentry{Individual trajectories}
    \addplot[atMean, line width=1.2pt, mark=*, mark size=1.3pt] table[x=turn, y=mean] {figures/agents_trajectories.dat};
    \addlegendentry{Average}
    % one session drawn solid, with the packages it requests at each turn:
    % cpu:agent_notebook[1432], column s3 (new_packages_per_turn.json)
    \addplot[atLine!70!black, line width=0.9pt, mark=*, mark size=1.1pt, forget plot]
      table[x=turn, y=s3] {figures/agents_trajectories.dat};
    \tikzset{at pk/.style={font=\scriptsize, text=atLine!60!black, fill=white, fill opacity=0.85, text opacity=1,
      inner sep=1pt, anchor=west}, at lead/.style={atLine!60!black, line width=0.3pt}}
    \draw[at lead] (axis cs:1,5) -- (axis cs:2.2,5.6) node[at pk] {numpy, pandas, matplotlib, seaborn, transformers};
    \draw[at lead] (axis cs:2,2) -- (axis cs:2.55,3.7) node[at pk] {+ sentence-transformers, umap-learn};
    \draw[at lead] (axis cs:3,1) -- (axis cs:3.35,2.55) node[at pk] {+ plotly};
    \draw[at lead] (axis cs:4,1) -- (axis cs:4.75,2.95) node[at pk] {+ scikit-learn};
  \end{axis}
\end{tikzpicture}%
\endgroup

%% file: figures/fig_layer_pull.tex
% Loading a container image (pandas, torch and nltk) with every layer pulled from the registry: panel (a) of the figure
% whose panel (b) is fig_import_phases.tex. The base and runtime layers may be cached on the host;
% an annotation gives what that saves (the ready time of the run with them cached, which the
% data file also holds, as \lpCacheSaving).
% Use: \input{figures/fig_layer_pull.tex} inside a figure; needs \usepackage{pgfplots}.
% Data: figures/layer_pull_data.tex, written by experiments/layer_pull/export_tex.py.
\begingroup
\input{figures/layer_pull_data.tex}%
\definecolor{lpDownload}{HTML}{2A78D6}%
\definecolor{lpExtract}{HTML}{EB6834}%
\definecolor{lpRun}{HTML}{6D6C67}%
\definecolor{lpCachedDot}{HTML}{B9B8B3}%
\definecolor{lpInk}{HTML}{52514E}%
% \lpPhase{bottom}{top}{start}{end}{color}, \lpNote{row}{x}{text}, \lpCached{row}
\def\lpPhase#1#2#3#4#5{\fill[#5, draw=white, line width=0.4pt] (axis cs:#3,#1) rectangle (axis cs:#4,#2);}%
\def\lpNote#1#2#3{\node[anchor=west, font=\scriptsize, inner sep=1.5pt] at (axis cs:#2,#1) {#3};}%
\def\lpKey#1#2{\textcolor{#1}{\rule[-0.15ex]{1.4ex}{1.4ex}}\,#2}%
\def\lpCached#1{\node[circle, fill=lpCachedDot, inner sep=0pt, minimum size=2.4pt] at (axis cs:0.15,#1) {};
  \node[anchor=west, font=\scriptsize, text=lpInk, inner sep=1.5pt] at (axis cs:0.4,#1) {cached on host, 0\,MB fetched};}%
\pgfplotsset{
  lp phases/.style={
    width=4.2cm, height=2.2cm, scale only axis,
    xmin=0, xmax=20, ymin=-0.6, ymax=3.6,
    ytick={0,...,3},
    yticklabels={Start + import, Custom layer (pip installs), Python runtime layer, Base layer (Debian)},
    yticklabel style={font=\scriptsize, align=right},
    ticklabel style={font=\scriptsize},
    xticklabel={\pgfmathprintnumber[assume math mode=true]{\tick}},
    y axis line style={draw=none}, ytick style={draw=none},
    axis x line*=bottom, x axis line style={lpInk}, xtick style={lpInk},
    xmajorgrids, grid style={black!10, line width=0.3pt},
    title style={font=\scriptsize, inner sep=2pt},
    clip=false,
  },
}%
\begin{subfigure}{\columnwidth}\centering
\begin{tikzpicture}
  \begin{axis}[lp phases, name=cold,
      xlabel={Time since pull request (s)}, xlabel style={font=\scriptsize}]
    \lpPhasesCold
    % the base and runtime layers are shared by many images and may already be on the host
    \draw[decorate, decoration={brace, amplitude=2.5pt}, thin, lpInk]
      (axis cs:9.2,3.35) -- (axis cs:9.2,1.65);
    \node[anchor=west, font=\scriptsize, text=lpInk, align=left, inner sep=1pt] at (axis cs:9.75,2.5)
      {May be cached on the host:\\saving \lpCacheSaving\,secs};
  \end{axis}
  \node[anchor=south, font=\scriptsize, inner sep=0pt, yshift=3pt, xshift=-8mm] at (cold.north) {%
    \lpKey{lpDownload}{Download}\hspace{0.8em}\lpKey{lpExtract}{Decompress + write}\hspace{0.8em}\lpKey{lpRun}{Start + import}};
\end{tikzpicture}%
\subcaption{Time to initialize and import packages in a custom image}\label{fig:layer_pull_a}
\end{subfigure}%
\endgroup

%% file: figures/fig_import_phases.tex
% Where time goes when a sandbox first imports a package: one grouped bar chart with a
% group per package and, in each group, a stacked bar per approach (from source, from
% __pycache__, by checkpoint restore).
% Panel (b) of the figure whose panel (a) is fig_layer_pull.tex.
% Use: \input{figures/fig_import_phases.tex} inside a figure; needs \usepackage{pgfplots}.
% Data: figures/import_phases_data.tex, written by experiments/import_phases/export_tex.py.
\begingroup
\input{figures/import_phases_data.tex}\definecolor{ipInterp}{HTML}{8A8984}%
\definecolor{ipSearch}{HTML}{2A78D6}%
\definecolor{ipFiles}{HTML}{1BAF7A}%
\definecolor{ipCompile}{HTML}{EB6834}%
\definecolor{ipBytecode}{HTML}{EDA100}%
\definecolor{ipMemory}{HTML}{6250D6}%
\definecolor{ipRestore}{HTML}{008300}%
\definecolor{ipInk}{HTML}{52514E}%
% The data file writes each package as \ipPanel{index}{name}{xmax}{segments}, with segments
% \ipSeg{bottom}{top}{start}{end}{color} on rows 2 (source), 1 (__pycache__) and 0 (restore),
% and totals \ipTotal{row}{time}{text}. Here a row becomes a bar: group i starts at x=4(i-1),
% and the bars run source, __pycache__, restore from left to right.
% Coordinates are expanded before \fill and \node reach the axis, because pgfplots runs them
% only at \end{axis}, when the macros hold the last package's values.
\def\ipPanel#1#2#3#4{%
  \def\ipSeg##1##2##3##4##5{%
    \pgfmathsetmacro\ipA{4*(#1-1)+2-##2}\pgfmathsetmacro\ipB{4*(#1-1)+2-##1}%
    \edef\ipDo{\noexpand\fill[##5, draw=white, line width=0.3pt]
      (axis cs:\ipA,##3) rectangle (axis cs:\ipB,##4);}\ipDo}%
  \def\ipTotal##1##2##3{%
    \pgfmathsetmacro\ipA{4*(#1-1)+2-##1}%
    \edef\ipDo{\noexpand\node[anchor=south, font=\noexpand\scriptsize, inner sep=1pt]
      at (axis cs:\ipA,##2) {\unexpanded{##3}};}\ipDo}%
  \pgfmathsetmacro\ipA{4*(#1-1)+1}%
  \edef\ipDo{\noexpand\node[anchor=north, font=\noexpand\scriptsize\noexpand\bfseries, inner sep=1pt]
    at (axis cs:\ipA,0) [yshift=-0.8cm] {#2};}\ipDo
  #4}%
\def\ipKey#1#2{\textcolor{#1}{\rule[-0.15ex]{1.4ex}{1.4ex}}\,#2}%
\begin{subfigure}{\columnwidth}\centering
\begin{tikzpicture}
  \begin{axis}[name=ip,
      width=7.15cm, height=1.8cm, scale only axis,
      xmin=-0.6, xmax=10.6, ymin=0, ymax=8,
      xtick={0,1,2,4,5,6,8,9,10},
      xticklabels={Source, Bytecode, Restore, Source, Bytecode, Restore, Source, Bytecode, Restore},
      xticklabel style={font=\scriptsize, text=black, rotate=45, anchor=east},
      yticklabel style={font=\scriptsize, text=black},
      yticklabel={\pgfmathprintnumber[assume math mode=true]{\tick}},
      ylabel={Time (s)}, ylabel style={font=\scriptsize, text=black},
      axis y line*=left, axis x line*=bottom, xtick style={draw=none},
      ymajorgrids, grid style={black!10, line width=0.3pt},
      clip=false]
    \ipPanels
  \end{axis}
  \node[anchor=south, font=\scriptsize, inner sep=0pt, yshift=0pt] at (ip.outer north) {%
    \begin{tabular}{@{}l@{\hspace{0.8em}}l@{\hspace{0.8em}}l@{}}
      \ipKey{ipInterp}{Load interpreter} & \ipKey{ipSearch}{Search \& resolve modules} & \ipKey{ipFiles}{Load files}\\
      \ipKey{ipCompile}{Compile to bytecode} & \ipKey{ipBytecode}{Load bytecode} & \ipKey{ipMemory}{Execute into memory}\\
      \ipKey{ipRestore}{Restore from snapshot} & & \\
    \end{tabular}};
\end{tikzpicture}%
\subcaption{Time to import a package across different methods}\label{fig:layer_pull_b}
\end{subfigure}%
\par\vspace{-6pt}% pull the figure caption closer to the sub-captions
\endgroup

%% file: figures/import_phases_data.tex
% Generated by experiments/import_phases/export_tex.py (median repetition). Do not edit.
\def\ipPanels{\ipPanel{1}{pandas}{3}{\ipSeg{1.69}{2.31}{0.0000}{0.0308}{ipInterp} \ipSeg{1.69}{2.31}{0.0308}{0.1313}{ipSearch} \ipSeg{1.69}{2.31}{0.1313}{0.2811}{ipFiles} \ipSeg{1.69}{2.31}{0.2811}{2.1835}{ipCompile} \ipSeg{1.69}{2.31}{2.1835}{2.7369}{ipMemory} \ipTotal{2}{2.7369}{2.74\,s} \ipSeg{0.69}{1.31}{0.0000}{0.0260}{ipInterp} \ipSeg{0.69}{1.31}{0.0260}{0.1000}{ipSearch} \ipSeg{0.69}{1.31}{0.1000}{0.4496}{ipBytecode} \ipSeg{0.69}{1.31}{0.4496}{1.0020}{ipMemory} \ipTotal{1}{1.0020}{1.00\,s} \ipSeg{-0.31}{0.31}{0.0000}{0.2912}{ipRestore} \ipTotal{0}{0.2912}{0.29\,s}}
\ipPanel{2}{torch}{6}{\ipSeg{1.69}{2.31}{0.0000}{0.0335}{ipInterp} \ipSeg{1.69}{2.31}{0.0335}{0.2670}{ipSearch} \ipSeg{1.69}{2.31}{0.2670}{0.6071}{ipFiles} \ipSeg{1.69}{2.31}{0.6071}{3.4544}{ipCompile} \ipSeg{1.69}{2.31}{3.4544}{5.1171}{ipMemory} \ipTotal{2}{5.1171}{5.12\,s} \ipSeg{0.69}{1.31}{0.0000}{0.0265}{ipInterp} \ipSeg{0.69}{1.31}{0.0265}{0.2598}{ipSearch} \ipSeg{0.69}{1.31}{0.2598}{0.7965}{ipBytecode} \ipSeg{0.69}{1.31}{0.7965}{2.7953}{ipMemory} \ipTotal{1}{2.7953}{2.80\,s} \ipSeg{-0.31}{0.31}{0.0000}{0.3081}{ipRestore} \ipTotal{0}{0.3081}{0.31\,s}}
\ipPanel{3}{nltk}{8}{\ipSeg{1.69}{2.31}{0.0000}{0.0313}{ipInterp} \ipSeg{1.69}{2.31}{0.0313}{0.4447}{ipSearch} \ipSeg{1.69}{2.31}{0.4447}{0.9185}{ipFiles} \ipSeg{1.69}{2.31}{0.9185}{5.8880}{ipCompile} \ipSeg{1.69}{2.31}{5.8880}{7.2809}{ipMemory} \ipTotal{2}{7.2809}{7.28\,s} \ipSeg{0.69}{1.31}{0.0000}{0.0267}{ipInterp} \ipSeg{0.69}{1.31}{0.0267}{0.3290}{ipSearch} \ipSeg{0.69}{1.31}{0.3290}{1.1538}{ipBytecode} \ipSeg{0.69}{1.31}{1.1538}{2.6792}{ipMemory} \ipTotal{1}{2.6792}{2.68\,s} \ipSeg{-0.31}{0.31}{0.0000}{0.4187}{ipRestore} \ipTotal{0}{0.4187}{0.42\,s}}}
\def\ipLast{3}

%% file: figures/fig_sharing.tex
% Per-set checkpoints (a) against one shared superset checkpoint (b), for the five most frequent
% Kaggle package sets. One column wide, (a) above (b); both read left to right.
% Use: \input{figures/fig_sharing.tex}; needs \usepackage{graphicx} and
% \usetikzlibrary{calc,fit,decorations.pathreplacing} in the preamble.
% Values: experiments/checkpoint_sharing/sharing.json ("example"). Every package block has the
% same width; sizes are the summed memory of the checkpoints' packages. In (b), each request restores
% a new container from the one shared checkpoint (a dot on the trunk). The container holds every
% package of the checkpoint: those the request uses in colour, the rest greyed; the package R4 and
% R5 import from disk (their residual) is the dashed block at the end of their container.
\begingroup
% The paper's palette at 0.85 strength, as the bars of fig_latency_hist.tex; package blocks are
% split by thin white rules, as the bars of fig_import_phases.tex.
\definecolor{shBlue}{HTML}{2A78D6}\definecolor{shOrange}{HTML}{EB6834}%
\definecolor{shGreen}{HTML}{1BAF7A}\definecolor{shYellow}{HTML}{EDA100}%
\definecolor{shPurple}{HTML}{8A5DB0}\definecolor{shGrey}{HTML}{B9B8B3}%
\def\shW{0.9}% cm, width of every package block
\tikzset{
  sh pk/.style={draw=white, line width=0.4pt, minimum height=3.6mm, minimum width=\shW cm, inner sep=0pt, outer sep=0pt,
    anchor=west, font=\scriptsize},
  sh np/.style={sh pk, fill=shBlue!85!white},
  sh pd/.style={sh pk, fill=shOrange!85!white},
  sh sk/.style={sh pk, fill=shGreen!85!white},
  sh mp/.style={sh pk, fill=shYellow!85!white},
  sh sn/.style={sh pk, fill=shPurple!85!white},
  sh xg/.style={sh pk, fill=shGrey!85!white},
  sh unused/.style={sh pk, fill=gray!6, draw=gray!55, text=gray!60},
  sh req/.style={font=\scriptsize, anchor=east},
  sh ttl/.style={font=\scriptsize\bfseries, anchor=west},
  sh ck/.style={draw, rounded corners=1.5pt, inner sep=1.2pt},
  sh ct/.style={draw, rounded corners=3pt, inner sep=1.8pt, line width=0.7pt},
  sh fork/.style={circle, fill=black, inner sep=0pt, minimum size=3.2pt},
  sh arr/.style={-latex, thin},
}%
% the two panels are separate pictures; (a) is padded to the width of (b), so that both start at the
% same x and their requests line up
\def\shRight{188.5pt}% right edge of (b), its "shared snapshot" label
\begin{subfigure}{\columnwidth}\centering
\begin{tikzpicture}[x=1cm,y=1cm]
  % ---------------- (a) one checkpoint per package set ----------------
  \foreach \i/\n/\pk in {
      1/2/{np/np,pd/pd},
      2/3/{np/np,pd/pd,sk/sk},
      3/4/{np/np,pd/pd,sk/sk,mp/mpl},
      4/5/{np/np,pd/pd,sk/sk,mp/mpl,sn/sns},
      5/4/{np/np,pd/pd,sk/sk,xg/xgb}} {
    \node[sh req] (shr\i) at (-0.25,-0.5*\i) {R\i};
    \coordinate (sha\i0) at (0.25,-0.5*\i);
    \foreach \s/\l [count=\j, evaluate=\j as \p using int(\j-1)] in \pk
      \node[sh \s] (sha\i\j) at (sha\i\p.east) {\l};
    \node[sh ck, fit=(sha\i1)(sha\i\n)] (shc\i) {};
    \draw[sh arr] (shr\i.east) -- (shc\i.west);
  }
  \path (-0.6,-0.5) -- (\shRight,-0.5);
\end{tikzpicture}%
\subcaption{One snapshot per request: 162\,MB total snapshot size}\label{fig:sharing_a}
\end{subfigure}\\[6pt]
\begin{subfigure}{\columnwidth}\centering
\begin{tikzpicture}[x=1cm,y=1cm]
  % ---------------- (b) one shared superset checkpoint ----------------
  \begin{scope}[yshift=-3.6cm]
    % the shared checkpoint, packages side by side as in (a), above the containers
    \coordinate (shb0) at (0.85,-0.3);
    \foreach \s/\l [count=\j, evaluate=\j as \p using int(\j-1)] in {np/np,pd/pd,sk/sk,mp/mpl}
      \node[sh \s] (shb\j) at (shb\p.east) {\l};
    \node[sh ck, fit=(shb1)(shb4)] (shshared) {};
    \node[font=\scriptsize, anchor=west] at ($(shshared.east)+(0.1,0)$) {shared snapshot};
    % each request restores the checkpoint into a new container (a dot on the trunk); the first \u
    % packages are the ones the request uses, the rest of the checkpoint is restored but unused
    \foreach \i/\u/\res/\resl in {1/2//, 2/3//, 3/4//, 4/4/sn/sns, 5/3/xg/xgb} {
      \pgfmathsetmacro\y{-0.95-0.56*(\i-1)}
      \node[sh req] (shq\i) at (-0.25,\y) {R\i};
      \node[sh fork] (shf\i) at (0.3,\y) {};
      \coordinate (shk\i0) at (0.85,\y);
      \foreach \s/\l [count=\j, evaluate=\j as \p using int(\j-1)] in {np/np,pd/pd,sk/sk,mp/mpl} {
        \ifnum\j>\u \node[sh unused] (shk\i\j) at (shk\i\p.east) {\l};
        \else \node[sh \s] (shk\i\j) at (shk\i\p.east) {\l}; \fi
      }
      \ifx\res\empty
        \node[sh ct, fit=(shk\i1)(shk\i4)] (shct\i) {};
      \else
        \node[sh \res, draw=black, dashed] (shx\i) at ($(shk\i4.east)+(0.1,0)$) {\resl};
        \node[sh ct, fit=(shk\i1)(shx\i)] (shct\i) {};
      \fi
      \draw (shq\i.east) -- (shf\i);
      \draw[sh arr] (shf\i) -- (shct\i.west);
    }
    % the trunk: the checkpoint is restored into every container
    \draw[semithick] (shshared.west) -| (shf5);
    % key
    \node[font=\scriptsize, anchor=west, inner sep=0pt] at (-0.55,-3.7) {%
      \tikz[baseline=-0.6ex]\node[sh fork]{};\,restore snapshot\hspace{0.45em}%
      \tikz[baseline=-0.6ex]\node[sh ct, minimum width=4mm, minimum height=2.6mm]{};\,container\hspace{0.45em}%
      \tikz[baseline=-0.6ex]\node[sh unused, minimum width=4mm, minimum height=2.6mm]{};\,unused\hspace{0.45em}%
      \tikz[baseline=-0.6ex]\node[sh pk, fill=white, draw=black, dashed, minimum width=4mm, minimum height=2.6mm]{};\,residual import};
  \end{scope}
  \path (-0.6,-4.1) -- (\shRight,-4.1);
\end{tikzpicture}%
\subcaption{A snapshot shared across requests: 32\,MB total snapshot size}\label{fig:sharing_b}
\end{subfigure}%
\par\vspace{-6pt}% pull the figure caption closer to the sub-captions
\endgroup

%% file: figures/fig_workload_clusters.tex
% Package sets of the three largest clusters of the training corpus, laid out by
% Jaccard distance (t-SNE), sized by request count and coloured by the checkpoint that serves them
% under the cost model. Each cluster has a
% checkpoint (filled circle with a thick outline, labelled c_1 to c_3
% with its packages) and an area outlining the cluster (shaded ellipse); the areas overlap where
% clusters meet. A key above tells requests from checkpoints.
% Use: \input{figures/fig_workload_clusters.tex} inside a figure; needs \usepackage{pgfplots}.
% Data: figures/workload_clusters.dat, workload_clusters_labels.tex and
% workload_clusters_areas.tex, written
% by experiments/workload_clusters/export_tex.py from clusters.py.
\begingroup
\input{figures/workload_clusters_labels.tex}%
\input{figures/workload_clusters_areas.tex}%
\definecolor{wcTab}{HTML}{2A78D6}%
\definecolor{wcVis}{HTML}{EB6834}%
\definecolor{wcNlp}{HTML}{1BAF7A}%
\definecolor{wcInk}{HTML}{52514E}%
% \wcLabel{x}{y}{label x}{label y}{anchor}{align}{text}: a leader line from the dot to its label
\def\wcLabel#1#2#3#4#5#6#7{\draw[wcInk, line width=0.25pt] (axis cs:#1,#2) -- (axis cs:#3,#4)
  node[anchor=#5, align=#6, font=\scriptsize, inner sep=1pt] {#7};}%
% \wcArea{colour}{path}: the area a checkpoint serves; \wcCkpt{colour}{n}{x}{y}: checkpoint c_n
\def\wcArea#1#2{\fill[#1, fill opacity=0.13] #2 -- cycle; \draw[#1!80!black, line width=0.5pt, dashed] #2 -- cycle;}%
% \wcCkpts runs twice: inside the axis it only records each checkpoint's position, and after the
% axis it draws the checkpoints there, so that they sit on top of every point
\def\wcCkptAt#1#2#3#4{\coordinate (wcck#2) at (axis cs:#3,#4);}%
\def\wcCkptDraw#1#2#3#4{%
  \node[circle, draw=black, line width=1.2pt, fill=#1, minimum size=8pt, inner sep=0pt] at (wcck#2) {};}%
\pgfplotsset{
  wc class/.style 2 args={only marks, mark=*, scatter, scatter src=explicit,
    visualization depends on={\thisrow{size} \as \wcSize},
    scatter/@pre marker code/.code={\scope[mark size=\wcSize pt, fill=#2, draw=none,
      line width=0.15pt, fill opacity=0.4]},
    scatter/@post marker code/.code={\endscope},
    x filter/.expression={\thisrow{class}==#1 ? x : nan}},
}%
\begin{tikzpicture}
  \begin{axis}[
      width=3.95cm, height=4.36cm, scale only axis, % same scale on both axes: the t-SNE x and y ranges are 32.2 and 35.0
      xmin=-0.4, xmax=10.4, ymin=-0.4, ymax=10.6,
      hide axis, unbounded coords=discard, clip=false,
    ]
    \wcAreas
    \addplot[wc class={0}{wcTab}] table[x=x, y=y, meta=size] {figures/workload_clusters.dat};
    \addplot[wc class={1}{wcVis}] table[x=x, y=y, meta=size] {figures/workload_clusters.dat};
    \addplot[wc class={2}{wcNlp}] table[x=x, y=y, meta=size] {figures/workload_clusters.dat};
    {\let\wcCkpt\wcCkptAt \wcCkpts}
    \wcLabels
  \end{axis}
  {\let\wcCkpt\wcCkptDraw \wcCkpts}
  % key, above the plot: requests are faded circles, checkpoints filled circles with a thick outline;
  % each label is anchored at its mid west, so its middle sits level with the centre of its symbol
  \coordinate (wckey) at ([yshift=16pt]current axis.north);
  \node[circle, fill=wcTab, fill opacity=0.4, minimum size=5pt, inner sep=0pt] (wck1) at ($(wckey)+(-1.25,0)$) {};
  \node[anchor=mid west, font=\scriptsize, inner sep=0pt] (wck1t) at ($(wck1.east)+(2pt,0)$) {request};
  \node[circle, draw=black, line width=1.2pt, fill=wcTab, minimum size=7pt, inner sep=0pt] (wck2)
    at ($(wck1t.east)+(0.45,0)$) {};
  \node[anchor=mid west, font=\scriptsize, inner sep=0pt] at ($(wck2.east)+(2pt,0)$) {snapshot};
\end{tikzpicture}%
\endgroup

%% file: sections/system_overview.tex
\section{System Overview}
\label{sec:system}

\begin{figure}[t]
    \centering
    \input{figures/fig_architecture.tex}

    \vspace{-2mm}
    \caption{Architecture of our system (\sys). 
        Our core contribution is to use a small number of learned snapshots.}
    \label{fig:readie_architecture}
\end{figure}
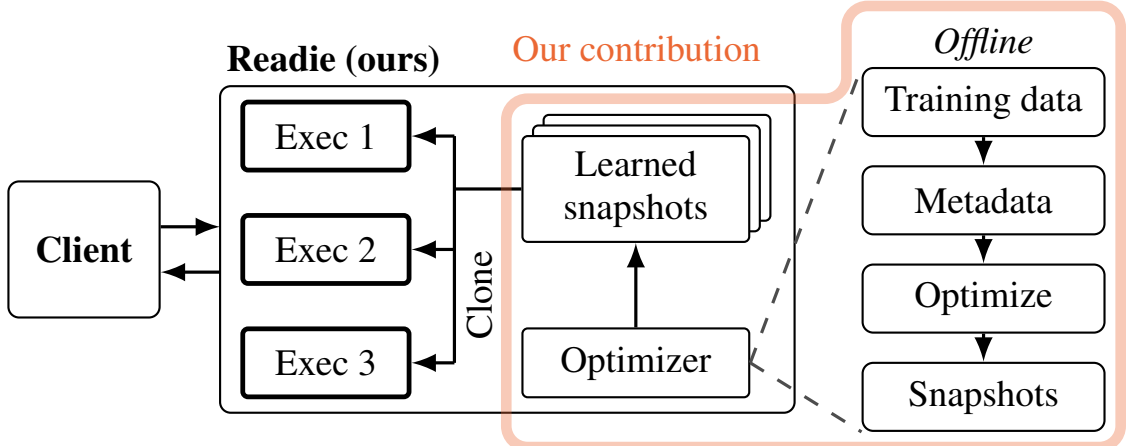

This section covers \sys's components and the path taken when a code execution is requested (\cref{fig:readie_architecture}).

\subsection{\sys Components}
\label{sec:system_components}
The client, router, and executors serve incoming requests, while the optimizer runs offline and produces the learned snapshots from which executors are cloned.
\, \emph{(Client)} The client sends the code to \sys and receives its response.
\, \emph{(Router)} The router selects a snapshot for the request based on the requirements of the code.
\, \emph{(Executor)} Each executor is a sandbox that runs the code. It is cloned from a learned snapshot (a checkpoint taken after importing the planned packages), so it starts with those packages already in memory.
\, \emph{(Optimizer)} Offline, the optimizer derives metadata (\cref{sec:implementation_implementation}) from historical requests (the training data), solves an optimization problem (\cref{sec:formulation_solving_optimization_problem}) to select a small set of snapshots, and builds and stores them on disk.

\subsection{\sys Workflow}
\label{sec:system_workflow}
A request goes through three steps, from the client to its execution on \sys (\cref{fig:readie_architecture}).
\, \emph{(Client analysis and serialization)} The client finds the modules the code imports and sends the serialized code and its arguments to the router.
\, \emph{(Snapshot selection and provisioning)} The router selects the snapshot that minimizes the request's initialization time (\cref{sec:formulation_initialization_time}), clones it into a new executor sandbox, and streams the payload to the executor over a socket.
\, \emph{(Code execution)} The executor deserializes the code and imports the needed modules; those held by the snapshot are already in memory, so only the rest are loaded from disk. It then runs the code and returns the serialized result (or exception) to the client along the same path.

%% file: figures/fig_architecture.tex
% The architecture in one figure, left to right: the client, the system box, and the
% offline pipeline drawn as a vertical flow. Inside the system box, three executors on
% the left are cloned from the learned snapshots on the right, which the
% optimizer below them produces; dashed lines join the optimizer to the offline flow.
% One column wide.
% Use: \input{figures/fig_architecture.tex}; needs \usetikzlibrary{calc,fit} in the
% preamble (already there for fig_sharing.tex). No data file.
\begingroup
\definecolor{saOurs}{HTML}{EB6834}%
\tikzset{
  sa tier/.style={draw, rounded corners=2pt},
  sa name/.style={font=\scriptsize\bfseries, inner sep=1pt},
  sa phase/.style={font=\scriptsize\itshape, inner sep=1pt},
  sa exec/.style={draw, line width=0.9pt, fill=white, rounded corners=1pt, font=\scriptsize,
    align=center, minimum width=1.1cm, minimum height=4.5mm, inner sep=2pt},
  sa part/.style={draw, fill=white, rounded corners=1pt, font=\scriptsize, align=center,
    minimum width=1.5cm, minimum height=4.5mm},
  sa ckpt/.style={draw, fill=white, rounded corners=1pt, font=\scriptsize, align=center,
    minimum width=1.5cm, minimum height=7mm, inner sep=1.5pt},
  sa stage/.style={draw, rounded corners=2pt, fill=white, font=\scriptsize,
    align=center, minimum width=1.6cm, minimum height=4.5mm, inner sep=1pt},
  sa arr/.style={-latex, semithick},
  sa lbl/.style={font=\scriptsize, align=center, inner sep=1.5pt},
}%
\begin{tikzpicture}[x=1cm,y=1cm]
  % ---------------- client ----------------
  \node[sa stage, minimum width=1.0cm, minimum height=9mm, font=\scriptsize\bfseries]
    (saclient) at (0.5,0) {Client};

  % ---------------- the system ----------------
  \draw[sa tier, fill=white] (1.4,-1.08) rectangle (5.2,1.08);
  \node[sa name, anchor=south west] at (1.4,1.1) {\sys (ours)};
  \foreach \i/\y in {1/0.75,2/0,3/-0.75}
    \node[sa exec] (saex\i) at (2.1,\y) {Exec \i};
  \foreach \k in {2,1}
    \node[sa ckpt] (saback\k) at ($(4.15,0.4)+(0.07*\k,0.07*\k)$) {};
  \node[sa ckpt] (sackpt) at (4.15,0.4) {Learned\\snapshots};
  \node[sa part] (saopt) at (4.15,-0.75) {Optimizer};
  \draw[sa arr] (saopt.north) -- (sackpt.south);
  % executors are cloned from the learned snapshots: a stem into a vertical bus
  \draw[semithick] (sackpt.west) -- (2.95,0.4);
  \draw[semithick] (2.95,0.75) -- (2.95,-0.75);
  \foreach \i in {1,2,3}
    \draw[sa arr] (saex\i -| 2.95,0) -- (saex\i.east);
  \node[sa lbl, rotate=90, anchor=north] at (2.97,-0.3) {Clone};

  % ---------------- request path, in above and back below ----------------
  \draw[sa arr] ($(saclient.east)+(0,0.15)$) -- (1.4,0.15);
  \draw[sa arr] (1.4,-0.15) -- ($(saclient.east)+(0,-0.15)$);

  % ---------------- offline optimization, a vertical flow to the right ----------------
  \foreach \i/\y/\txt in {1/0.975/{Training data},2/0.325/{Metadata},
                          3/-0.325/{Optimize},4/-0.975/{Snapshots}}
    \node[sa stage] (saoff\i) at (6.45,\y) {\txt};
  \foreach \i/\j in {1/2,2/3,3/4}
    \draw[sa arr] (saoff\i) -- (saoff\j);
  \node[sa phase, anchor=south] (saofflbl) at (saoff1.north) {Offline};

  % our contribution: a stepped outline hugging the snapshots, the optimizer and the offline flow
  \def\saPad{0.1}%
  \draw[saOurs, draw opacity=0.35, line width=2pt, rounded corners=4pt]
    ($(sackpt.west |- saoff4.south)+(-\saPad,-\saPad)$)
    -- ($(sackpt.west |- saback2.north)+(-\saPad,\saPad)$)
    -- ($(saoff1.west |- saback2.north)+(-\saPad,\saPad)$)
    -- ($(saoff1.west |- saofflbl.north)+(-\saPad,\saPad)$)
    -- ($(saoff1.east |- saofflbl.north)+(\saPad,\saPad)$)
    -- ($(saoff4.south east)+(\saPad,-\saPad)$)
    -- cycle;
  \node[font=\scriptsize, text=saOurs, anchor=south west, inner sep=1pt]
    at ([xshift=-\saPad cm]sackpt.west |- 0,1.2) {Our contribution};

  % the offline pipeline is what the optimizer runs
  \draw[dashed, semithick, black!70] (saopt.east) -- (saoff1.north west);
  \draw[dashed, semithick, black!70] (saopt.east) -- (saoff4.south west);
\end{tikzpicture}%
\endgroup

%% file: sections/formulation.tex
\section{Determining Ideal Snapshots}
\label{sec:formulation}

This section discusses how \sys determines which packages to include in each snapshot, how many snapshots to build, and which snapshot to restore for each request. The objective is to find snapshot sets which minimize the overall initialization time of all requests, while staying within a storage budget~\cite{sieve,sc_materialization}.

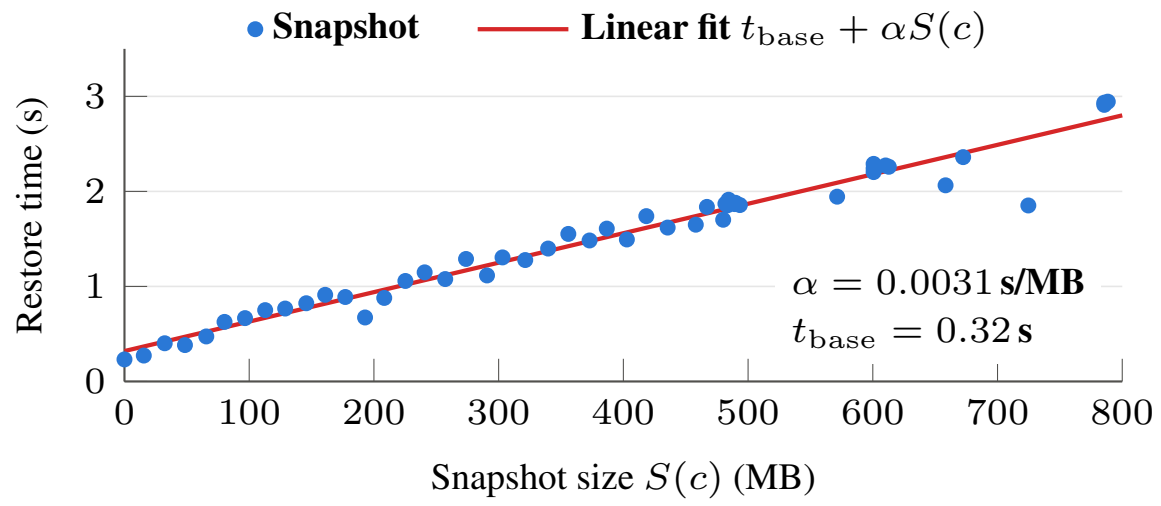
\begin{figure}[t]
    \centering
    \input{figures/fig_alpha.tex}
    \vspace{-2mm}
    \caption{Restore time against snapshot size for 50 snapshots, each holding a different set of packages. Each point is the median of 5 restores. The slope of the fitted line is $\alpha$ and its intercept is $t_{\mathrm{base}}$.}
    \label{fig:alpha_measurement}
\end{figure}

\subsection{Serving a Request with a Snapshot}
\label{sec:formulation_definitions}

\paragraph{Packages and their properties.}
Let $P$ be the union of all packages requested in the historical requests. Each package $p \in P$ has three properties that are required for planning snapshots: its dependencies, import time from disk ($t_p$), and size in memory ($m_p$). The process of measuring each of these properties is explained in \cref{sec:implementation_implementation}. For a set of packages $X$, we write $\overline{X}$ for its \emph{closure}, i.e., $X$ together with all its transitive
dependencies.

\paragraph{Request.}
A request is a function call, where $r$ is a set of packages it imports. Importing $r$ loads its closure, so in practice the request needs the packages $D_r = \overline{r}$.

\paragraph{Snapshot.}
A snapshot represents a sandbox which was checkpointed after importing a set of packages $c$~\cite{elasticnotebook,elasticnotebook_demo}. Again, importing $c$ loads its closure, so in practice the snapshot contains the packages $D_c = \overline{c}$. The size of a snapshot is the total memory that the packages $D_c$ hold, $S(c) = \sum_{p \in D_c} m_p$.

\subsection{Initialization Time}
\label{sec:formulation_initialization_time}

\subsubsection{Restore Time for a Snapshot}
Restoring a snapshot takes (1) a constant time $t_{\mathrm{base}}$ to create the sandbox, plus (2) time to restore its memory, which grows with its size: $T_{\mathrm{restore}}(c) = t_{\mathrm{base}} + \alpha \, S(c)$~\cite{airindex}, where $\alpha$ is the restore time in seconds per megabyte of memory and represents how the snapshot's memory state restoration time changes with its size. 

To test our modeling of the restore time, we restore 50 snapshots, each holding a different set of packages, and time each restore (\cref{fig:alpha_measurement}). The slope of the fitted line is $\alpha$, and the Y-intercept is $t_{\mathrm{base}}$. In our experiment, the fitted line has slope $\alpha = 0.0031$\,s/MB and intercept $t_{\mathrm{base}} = 0.32$\,s, which is consistent with the restore time having a constant component and a component that grows linearly with the snapshot size.

\subsubsection{Residual Import Time}
The residual import time accounts for the import time of the packages that a request $r$ needs and a snapshot $c$ lacks. Since package dependencies are not mutually exclusive, it is possible that the snapshot closure $D_c$ already contains some of the dependencies from the request closure $D_r$. The residual import time is then: $T_{\mathrm{residual}}(r, c) = \sum_{p \in D_r \setminus D_c} t_p$.

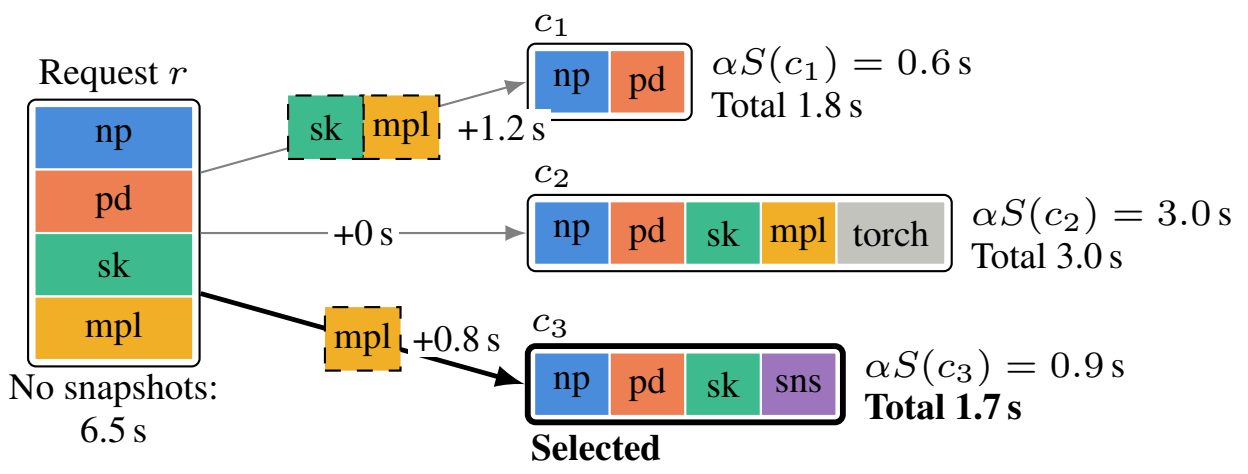
\begin{figure}[t]
    \centering
    \input{figures/fig_checkpoint_selection.tex}
    \caption{Selecting a snapshot for a request $r$. Beside each snapshot are its restore time $\alpha S(c)$ and its total time. Each arrow shows the residual packages and their import time. Snapshot $c_2$ needs no residual imports, but it holds PyTorch, which $r$ does not use, and is the slowest to restore. Snapshot $c_3$ lacks Matplotlib yet has the lowest total time, so it is selected.}
    \label{fig:snapshot_selection}
\end{figure}

Thus, the total initialization time of a request $r$ served from snapshot $c$ is the restore time plus the residual import time. $T(r, c) = t_{\mathrm{base}} + \alpha \, \sum_{p \in D_c} m_p + \sum_{p \in D_r \setminus D_c} t_p$. The two terms in this equation pull in opposite directions. A snapshot that holds more packages leaves a smaller residual but takes more time to restore, including for the packages the request does not use. With a sparse snapshot, the residual time increases as more packages are imported from disk. When a new request $r$ arrives, the snapshot $c$ with the lowest initialization time ($T(r, c)$) among all planned snapshots is selected to serve the request, as shown in \cref{fig:snapshot_selection}. $t_{\mathrm{base}}$ is left out since it is the same for every snapshot.

\subsection{Defining the Optimization Problem}
\label{sec:formulation_defining_optimization_problem}

The objective is to select a set of snapshots $C$ subject to the storage budget $B$ (MB), which minimizes the total initialization time on a set of requests $R$, when each request in $R$ is served by exactly one of the snapshots in $C$:
\begin{equation}
  \mathcal{F}(C) = \sum_{r \in R} \; \min_{c \in C} T(r, c) \qquad s.t. \quad \sum_{c \in C} S(c) \le B
  \label{eq:objective_function}
\end{equation}

% \paragraph{Constraints.}
% \begin{itemize}
%   \item \textbf{Storage.} The total size of all snapshots in $C$ should not exceed the budget of $B$ megabytes.
%   \item \textbf{One snapshot per request.} Every request is served by exactly one snapshot.
% \end{itemize}

\paragraph{Assumptions.}
The cost model only considers the packages present in the planning set and assumes that they are available on disk, treating each package's size in memory and import time as additive once its dependencies are loaded.

\subsection{Solving the Optimization Problem}
\label{sec:formulation_solving_optimization_problem}

The space of possible snapshots, which is every closed set of packages, grows exponentially with the number of packages, making the problem $\mathsf{NP}$-hard~\cite{kariv1979pmedian}. To limit the candidate space, we restrict the snapshot candidates to the distinct closures of the planning requests.

By limiting the candidate space, \cref{eq:objective_function} becomes a budgeted facility location problem where the distinct request closures are the customers as well as candidates for the facilities (snapshots). The distance from a customer $r$ to a facility $c$ is $T(r, c)$, which is the asymmetric distance that we want to minimize. The cost to open a facility is $S(c)$, and the total cost of all open facilities must not exceed the budget $B$. Every customer (request) is assigned to an open facility (snapshot). \sys solves it with a greedy algorithm for facility location~\cite{cornuejols1977location} which, at each step, adds the candidate that fits in the remaining budget and reduces the weighted total of \cref{eq:objective_function} the most. It stops when no candidate fits the budget or lowers the total. The time complexity of the algorithm is $O(K N^2)$ with a space complexity of $O(N^2 + NP)$, where $N$ is the number of candidates (here requests), $P$ the number of packages and $K$ the number of snapshots chosen.

% \begin{figure}[t]
%     \centering
%     \input{figures/fig_facility_location.tex}
%     \caption{Snapshot planning as budgeted facility location. Every request is a customer and a candidate facility. Each edge from a request to a snapshot represents the request's initialization time $T(r, c)$ using that snapshot, and its color shows the snapshot that serves the request: the nearest one, with the lowest $T(r, c)$. The snapshots' sizes $S(c)$ together fit in the storage budget $B$.}
%     \label{fig:snapshot_as_facility_location}
% \end{figure}

\paragraph{Alternative algorithm considerations.}
The budgeted facility location \textbf{MILP} provides an optimal solution, but its solve time grows exponentially with the number of distinct sets and packages in the worst case. \textbf{Greedy knapsack} ranks candidates once by the import time they save and takes them while they fit in the budget, ignoring what earlier picks already cover. \textbf{$k$-medoids}~\cite{kaufman1990pam} starts with the greedy facility location solution and a few random starts and refines it by swapping and adding snapshots under the budget, so it is never worse than greedy facility location itself. \textbf{Agglomerative Jaccard}~\cite{jaccard1901etude} clusters the request sets by Jaccard distance with average linkage and marks the union of each cluster as a snapshot, ignoring import time and size. \cref{sec:evaluation_optimal_snapshot} evaluates all the above algorithms on the planning and validation sets.

%% file: figures/fig_alpha.tex
% Restore time against checkpoint size: one point per checkpoint (the median of its timed
% restores) and the least-squares line through them. The slope is alpha and the intercept
% is t_base. One column wide; same axis style as fig_planners.tex.
% Use: \input{figures/fig_alpha.tex}; needs \usepackage{pgfplots}.
% Data: alpha_restores.dat and alpha_data.tex, from experiments/alpha/export_tex.py.
\begingroup
\input{figures/alpha_data.tex}%
\definecolor{alPoint}{HTML}{2A78D6}%
\definecolor{alFit}{HTML}{D62728}%
\definecolor{alInk}{HTML}{52514E}%
\begin{tikzpicture}
  \begin{axis}[
      width=6.6cm, height=2.2cm, scale only axis,
      xmin=0, xmax=\alXmax, ymin=0, ymax=\alYmax,
      xtick distance=100, ytick distance=1,
      ticklabel style={font=\scriptsize}, label style={font=\scriptsize},
      xlabel={Snapshot size $S(c)$ (MB)}, ylabel={Restore time (s)},
      axis lines*=left, axis line style={alInk}, tick style={alInk},
      ymajorgrids, grid style={black!10, line width=0.3pt},
      legend style={font=\scriptsize\bfseries, draw=none, fill=none, at={(0.5,0.97)}, anchor=south, legend columns=2, inner sep=0pt,
        /tikz/every even column/.append style={column sep=0.3cm}},
      legend cell align=left, clip=false,
    ]
    \addplot[only marks, mark=*, mark size=1.3pt, alPoint] table[x=size_mb, y=median_s] {figures/alpha_restores.dat};
    \addplot[alFit, line width=0.8pt, domain=0:\alXmax, samples=2] {\alBase + \alAlpha*x};
    \legend{Snapshot, Linear fit $t_{\mathrm{base}} + \alpha S(c)$}
    \node[anchor=south east, align=left, font=\scriptsize\bfseries, fill=white, inner sep=3pt]
      at (rel axis cs:0.98,0.05)
      {$\alpha = \alAlpha$\,s/MB\\[1pt]$t_{\mathrm{base}} = \alBase$\,s};
  \end{axis}
\end{tikzpicture}%
\endgroup

%% file: figures/fig_checkpoint_selection.tex
% Selecting a checkpoint for one request. The request is on the left and the checkpoints on the
% right, each with its restore term alpha*S(c) and its total beside it. An arrow goes from the
% request to each checkpoint, carrying the packages still to import after that checkpoint is
% restored (the residual, dashed) and their import time. The selected checkpoint and its arrow are
% drawn heavier. One column wide; package blocks and dashed residual imports as in
% fig_sharing.tex. Illustrative values, not measurements: the import times are rounded from the
% pipeline metadata (numpy 1.6 s, pandas 3.7 s, scikit-learn 0.4 s, Matplotlib 0.8 s), and the
% restore terms are made up to show the trade-off.
% Use: \input{figures/fig_checkpoint_selection.tex}; needs \usetikzlibrary{calc,fit}.
\begingroup
% The paper's palette at 0.85 strength, as the bars of fig_latency_hist.tex; package blocks are
% split by thin white rules, as the bars of fig_import_phases.tex.
\definecolor{csBlue}{HTML}{2A78D6}\definecolor{csOrange}{HTML}{EB6834}%
\definecolor{csGreen}{HTML}{1BAF7A}\definecolor{csYellow}{HTML}{EDA100}%
\definecolor{csPurple}{HTML}{8A5DB0}\definecolor{csGrey}{HTML}{B9B8B3}%
\tikzset{
  cs pk/.style={draw=white, line width=0.4pt, minimum height=4.2mm, minimum width=0.5cm, inner sep=0pt, outer sep=0pt,
    anchor=west, font=\scriptsize},
  cs np/.style={cs pk, fill=csBlue!85!white},
  cs pd/.style={cs pk, fill=csOrange!85!white},
  cs sk/.style={cs pk, fill=csGreen!85!white},
  cs mp/.style={cs pk, fill=csYellow!85!white},
  cs sn/.style={cs pk, fill=csPurple!85!white},
  cs tc/.style={cs pk, fill=csGrey!85!white, minimum width=0.72cm},
  cs res/.style={draw=black, dashed},
  cs ck/.style={draw, rounded corners=1.5pt, inner sep=1.2pt},
  cs arr/.style={-latex, black!50, thin},
  cs pick/.style={-latex, black, line width=0.9pt},
  cs lbl/.style={font=\scriptsize, inner sep=1pt},
}%
% #1 row y, #2 name, #3 held blocks (style/label), #4 index of the last block,
% #5 alpha*S label, #6 total label, #7 extra frame style
\newcommand\csCkpt[7]{%
  \coordinate (cs@0) at (3.38,#1);
  \foreach \s/\l [count=\j, evaluate=\j as \p using int(\j-1)] in {#3}
    \node[cs \s] (cs@\j) at (cs@\p.east) {\l};
  \node[cs ck, #7, fit=(cs@1)(cs@#4)] (#2) {};
  \node[cs lbl, anchor=south west] at (#2.north west) {$#2$};
  \node[cs lbl, anchor=west, align=left] at ($(#2.east |- 0,#1)+(0.08,0)$) {#5\\#6};
}%
% #1 checkpoint node, #2 start offset on the request, #3 residual blocks (style/label, may be empty),
% #4 residual time, #5 arrow style
\newcommand\csArrow[5]{%
  \draw[#5] ($(csreq.east)+(0,#2)$) -- (#1.west) coordinate[pos=0.5] (csmid);
  \coordinate (csr@0) at ($(csmid)+(-0.25*\csN,0)$);
  \foreach \s/\l [count=\j, evaluate=\j as \p using int(\j-1)] in {#3}
    \node[cs \s, cs res] (csr@\j) at (csr@\p.east) {\l};
  \ifnum\csN=0 \def\csAnc{center}\else\def\csAnc{west}\fi
  \node[cs lbl, anchor=\csAnc, fill=white] at ($(csmid)+(0.25*\csN+0.04*\csN,0)$) {#4};
}%
\begin{tikzpicture}[x=1cm,y=1cm]
  % ---------------- the request ----------------
  \node[cs np, anchor=north, minimum width=1.05cm] (csq1) at (0.6,0.84) {np};
  \node[cs pd, anchor=north, minimum width=1.05cm] (csq2) at (csq1.south) {pd};
  \node[cs sk, anchor=north, minimum width=1.05cm] (csq3) at (csq2.south) {sk};
  \node[cs mp, anchor=north, minimum width=1.05cm] (csq4) at (csq3.south) {mpl};
  \node[cs ck, fit=(csq1)(csq4)] (csreq) {};
  \node[cs lbl, anchor=south] at (csreq.north) {\scriptsize Request $r$};
  \node[cs lbl, anchor=north, align=center] at (csreq.south) {No snapshots:\\6.5\,s};

  % ---------------- the checkpoints ----------------
  \csCkpt{1.0}{c_1}{np/np, pd/pd}{2}{$\alpha S(c_1) = 0.6$\,s}{Total 1.8\,s}{}
  \csCkpt{0.0}{c_2}{np/np, pd/pd, sk/sk, mp/mpl, tc/torch}{5}{$\alpha S(c_2) = 3.0$\,s}{Total 3.0\,s}{}
  \csCkpt{-1.0}{c_3}{np/np, pd/pd, sk/sk, sn/sns}{4}{$\alpha S(c_3) = 0.9$\,s}{\textbf{Total 1.7\,s}}{line width=1.1pt}

  % ---------------- the residual imports along each arrow ----------------
  \def\csN{2}\csArrow{c_1}{0.4}{sk/sk, mp/mpl}{+1.2\,s}{cs arr}
  \def\csN{0}\csArrow{c_2}{0}{}{+0\,s}{cs arr}
  \def\csN{1}\csArrow{c_3}{-0.4}{mp/mpl}{+0.8\,s}{cs pick}
  \node[cs lbl, font=\scriptsize\bfseries, anchor=north west] at ($(c_3.south west)+(0,-0.02)$) {Selected};
\end{tikzpicture}%
\endgroup

%% file: sections/implementation.tex
\section{Implementation and Discussion}
\label{sec:implementation}

This section describes \sys's implementation (\cref{sec:implementation_implementation}) and design considerations (\cref{sec:implementation_discussion}).

\subsection{Implementation}
\label{sec:implementation_implementation}

\paragraph{Metadata generation.}
For each package, \sys imports it in a fresh interpreter and records the modules the import adds to \texttt{sys.modules}, which gives its dependencies and submodules. In a second fresh interpreter, it imports the dependencies first and then imports the package itself. The import time is the time this import takes, and the memory size is the growth of the interpreter's anonymous resident memory (\texttt{RssAnon}). Importing the dependencies first makes the measurements additive, so no package's memory or import time is counted more than once. The page cache is cleared before every run, so cached pages cannot shorten the import time.

\paragraph{Integrating with Python.}
A \texttt{@remote} decorator serializes the function and its arguments with cloudpickle~\cite{cloudpickle,chipmink,transactional_python}, finds the modules it imports from the code's abstract syntax tree, and sends both to the router and deserializes the response received after the execution.

\paragraph{Isolation.}
Each sandbox is a gVisor~\cite{gvisor} container with its own network stack, which places a user-space kernel between the code and the host for security and isolation.

\paragraph{Sessions.}
\sys keeps containers warm with a TTL so that requests with the same session ID go to the same sandbox and are able to reuse its imported modules and share global variables across executions~\cite{kishu,kishu_demo,codedata_versioning}.

\subsection{Design Considerations}
\label{sec:implementation_discussion}

\paragraph{Using a fat image for the rootfs.}
The root file system of every sandbox is a copy-on-write overlay, held in the sandbox's memory, over a read-only root file system which is the \citet{kaggle-docker-python} image that all sandboxes share.

\paragraph{File system support.}
A sandbox does not share the host's file system. This isolates users from each other and from the host, and keeps the root file system identical to the one the snapshots were checkpointed on, which a restore requires. File data can still be used by loading it into memory as a variable and passing it to the remote function, or it can be reused across multiple executions using sessions.

\paragraph{Packages with import side effects.}
The cost model (\cref{sec:formulation_initialization_time}) assumes that the packages in $D_c \setminus D_r$ do not affect a request's execution. This does not hold for packages that modify other modules on import, such as by monkey-patching. A request served from a snapshot that contains one executes with the modified modules even if the request does not import it. Such packages can be excluded from the snapshots in the planning stage and imported after restore when a request requires them.

% \paragraph{Integration with existing providers.}
% \sys's optimizer and router which are responsible for the planning of snapshots and selecting a snapshot based on the request respectively, are independent of the sandbox runtime. Hence, these components can be easily decoupled and used along with the snapshot creation and execution APIs of services such as Lambda
% SnapStart \cite{aws_lambda_snapstart} or Modal's memory snapshots \cite{modal_memory_snapshots}.

% \paragraph{Extending snapshots beyond packages to datasets and models.}
% The same cost model (\cref{sec:formulation_initialization_time}) can be applied to datasets and models, since it only ever requires the dependencies, import time and memory size. The import time in this case is the download time and time taken to load the dataset/model into a Python variable. Since its a variable, its memory size can be measured the same way as it is done for imports. Finally, their dependencies are basically the packages that are used to load the dataset/model itself.

%% file: sections/experiments.tex
\section{Experimental Evaluation}
\label{sec:evaluation}
In this section, we evaluate \sys on a variety of data science workloads, use it as a sandbox for agents on different agent benchmarks, and compare it with existing sandbox provisioning methods. We make the following claims:

\begin{enumerate}
  \item \textbf{Low initialization time:} At the same storage budget of 15\,GB, \sys initializes a sandbox 3$\times$ faster than using per-request snapshots with caching and 6.3$\times$ faster than using a fat image (\cref{sec:evaluation_low_init}).
  \item \textbf{Per-request latency at a fraction of the storage:} \sys matches the initialization time of using per-request snapshots, with 14$\times$ less storage (\cref{sec:evaluation_low_storage}).
  \item \textbf{Robust to unseen workloads:} In an out-of-distribution set, generated without knowledge of the requests \sys plans on, \sys still initializes 2$\times$ faster than its per-request snapshots (\cref{sec:evaluation_random_starts}).
  \item \textbf{Faster agent sandbox execution:} In agent sandbox executions tested on different agent benchmarks, \sys cuts initialization time by 9.1$\times$ and end-to-end time by 1.3$\times$ against each benchmark's provided sandbox environment (\cref{sec:evaluation_agent}).
  \item \textbf{Optimized for data science workflows:} In each of the 8 domain groups, \sys's initialization times are 4.6--5.6\% of the cold start time, of which residual imports account for only 9.8\% of the initialization time (\cref{sec:evaluation_optimized_for_ds}).
  \item \textbf{Near-optimal snapshot sets in less time:} The total latency of serving requests calculated from the output of the greedy facility location algorithm used by \sys is 22$\times$ lower, compared to the 12.7$\times$ reduction by agglomerative Jaccard clustering against having no snapshots at all (\cref{sec:evaluation_optimal_snapshot}).
\end{enumerate}

%%% experiments %%%
\subfile{./experiments/setup}
\subfile{./experiments/init}

\subfile{./experiments/snapshot}
\subfile{./experiments/unseen}
\subfile{./experiments/agents}
\subfile{./experiments/optimized_ds}

\subfile{./experiments/algorithm}
%%% end experiments %%%

%% file: sections/experiments/setup.tex
\subsection{Experiment Setup}
\label{sec:evaluation_experiment_setup}

\paragraph{Data.}
We have constructed our dataset using requests from seven sources (\cref{table:source_distribution}): code snippets and notebooks generated by LLMs, sessions in which an agent debugs failing code, and snippets and notebooks from \citet{kaggle} and \citet{huggingface}. They span 16 data science domains, such as exploratory data analysis, machine learning, computer vision, natural language processing, etc., each having an equal weight.
\, \emph{(Planning set)} Only the planning set is used by \sys to plan the ideal snapshots.
\, \emph{(Validation set)} The validation set is used to compare algorithms (\cref{sec:evaluation_optimal_snapshot}) and to choose the storage budget for snapshots, past which more storage brings little further reduction in initialization time.
\, \emph{(Test set)} This is a held-out set from the same distribution as the planning and validation sets on which evaluations are run.
\, \emph{(Out-of-distribution set)} Another set used for evaluations with the difference being that it is generated by a different agent with random sources and no knowledge of the requests in the other sets.

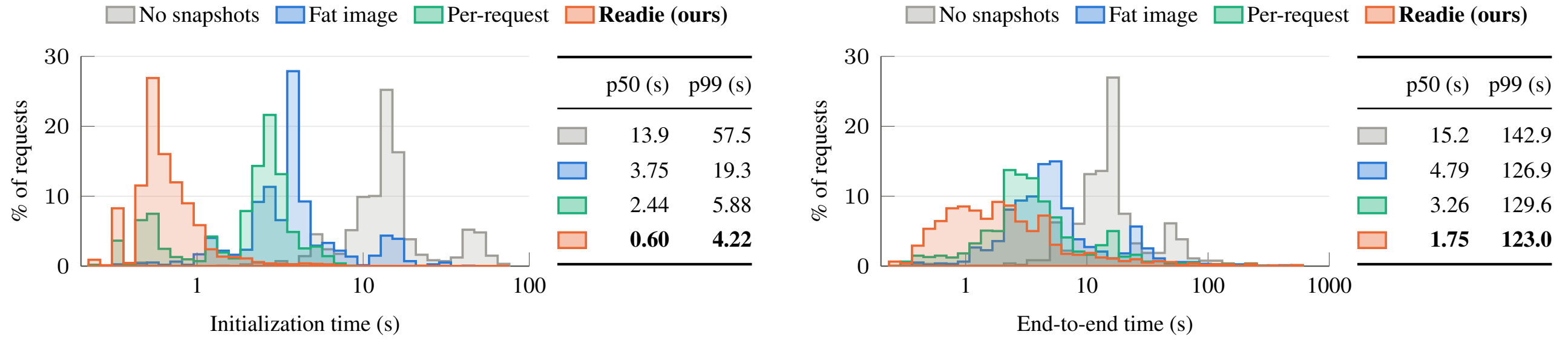
\begin{figure*}[t]
  \centering
  \def\ohSet{test}\input{figures/fig_methods_dist.tex}

  \vspace{-1mm}
  \caption{Distribution of (a) the initialization time and (b) the end-to-end time of the 1,535 requests in the test set, with every method under the same 15\,GB storage budget. Each histogram shows the percentage of requests in each time bin on a log scale, and the table beside it gives each method's p50 and p99. Most requests initialize in under a second with \sys, against a few seconds with the fat image and per-request snapshots and over ten seconds with no snapshots, and the gap carries over to the end-to-end time.}
  \label{fig:test_methods}
\end{figure*}

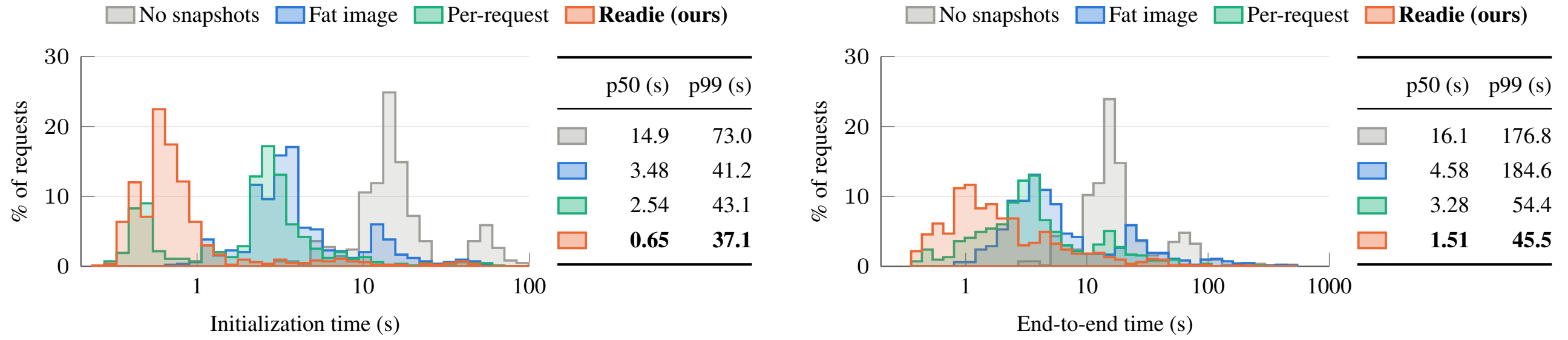
\begin{figure*}[t]
  \centering
  \def\ohSet{agent}\input{figures/fig_methods_dist.tex}

  \vspace{-1mm}
  \caption{Distribution of (a) the initialization time and (b) the end-to-end time of the 832 requests in the out-of-distribution set, which \sys did not see when it planned its snapshots. Each histogram shows the percentage of requests in each time bin on a log scale, and the table beside it gives each method's p50 and p99. Per-request snapshots are an optimistic, hypothetical baseline here, as each unseen request is given its own snapshot. \sys still initializes most requests in under a second and has the lowest p50 and p99 for both metrics.}
  \label{fig:agent_methods}
\end{figure*}

\begin{table}[t]
\caption{Count of requests from each source in each dataset. The out-of-distribution set is randomly generated and does not contain any source information.}
\label{table:source_distribution}

% \vspace{-2mm}

\centering
\small
\begin{sc}
\setlength{\tabcolsep}{4pt}
\begin{tabular}{lrrr}
\toprule
Source & Plan & Val & Test \\
\midrule
LLM generated snippets           & 2,890 & 110 & 601 \\
LLM generated notebooks          & 2,794 & 218 & 457 \\
Agent debugging sessions         & 2,947 & 531 & 177 \\
Kaggle snippets                  & 2,108 & 332 & 140 \\
Kaggle notebooks                 & 3,350 & 492 & 145 \\
Hugging Face snippets             &   722 &  68 &  28 \\
Hugging Face notebooks            &   488 &  70 &  11 \\
\midrule
Total                  & 15,299 & 1,821 & 1,559 \\
\bottomrule
\end{tabular}
\end{sc}
% \vskip -0.1in
\end{table}

\paragraph{Metrics.}
Each request runs in a fresh sandbox. A request's \emph{initialization time} is the time from provisioning until its sandbox has loaded every module the request imports. Without a snapshot, it is the total time to provision the sandbox, start the interpreter and import every module from disk, after first installing any missing packages, which is also the \emph{cold start time}. With a snapshot, it is the time to provision the sandbox, fetch the snapshot from a local disk or remote cache, restore the snapshot, and install and import the residual modules the snapshot lacks. The \emph{end-to-end time} is the initialization time plus the time to run the request's code, step by step, in that sandbox.

\paragraph{Methods.}
We compare \sys with three commonly used methods to initialize a sandbox:
\begin{itemize}
  \item \textbf{No snapshots:} The environment is built on demand, as installing a request's dependencies at startup. Every sandbox starts from a \texttt{python:3.12} image, which has Python and no installed packages. When a request arrives, the required packages are installed using \texttt{pip} in its sandbox and then imported. Nothing is stored for a request, and every request pays both installation and import.
  \item \textbf{Fat image:} One large image with the packages of almost every request pre-installed, which every sandbox shares, as notebook services do. We use the Kaggle Python 3.12 image~\cite{kaggle-docker-python}. A request does not install anything, but its new interpreter still imports every package it needs from disk.
  \item \textbf{Per-request snapshots:} Each snapshot is a sandbox that has imported exactly one distinct set of modules, so a request restores the snapshot of its own set and has nothing left to import. Covering the test and out-of-distribution sets takes 1,308 snapshots and 279\,GB.
  \item \textbf{\sys (ours):} A set of snapshots planned by our greedy facility location algorithm (\cref{sec:formulation_solving_optimization_problem}) on the planning set and at runtime the router picks the snapshot with the lowest $T(r, c)$ to serve each request and imports the residual.
\end{itemize}

\paragraph{Environment.}
Experiments run on an Ubuntu server with 2 AMD EPYC 7552 48-core processors and 1\,TB RAM. Each sandbox is run using gVisor and is provided with 8\,GB RAM and runs on its own 8 cores.

%% file: figures/fig_methods_dist.tex
% Init (left) and end-to-end (right) time distributions, two subfigures, each one chart that overlays
% the four methods' histograms, with a table of each method's p50 and p99 on its right.
% Use: \begin{figure*}[t]\centering\def\ohSet{test}\input{figures/fig_methods_dist.tex}\caption{...}\end{figure*},
% with \ohSet test or agent (the out-of-distribution set); labels fig:<set>_methods_{init,e2e};
% needs \usepackage{pgfplots}, \usepackage{subcaption}, \usepackage{booktabs} and \usetikzlibrary{calc}.
% Data: figures/hist_<set>_<metric>.dat and figures/latency_hist_data.tex, written by experiments/methods/export_tex.py.
\begingroup
\input{figures/latency_hist_data.tex}\definecolor{ohnockpt}{HTML}{9E9D98}% gray
\definecolor{ohfat}{HTML}{2A78D6}% blue
\definecolor{ohprmin}{HTML}{1BAF7A}% green
\definecolor{ohbounded}{HTML}{EB6834}% orange, ours
\definecolor{ohInk}{HTML}{52514E}%
\def\ohv#1#2{\csname lh-\ohSet-\ohMetric-#1-#2\endcsname}%
\pgfplotsset{
  oh panel/.style={
    width=4.9cm, scale only axis,
    xmin=0.2, xtick={1,10,100,1000}, xticklabels={1,10,100,1000}, ymin=0, ymax=30, ytick={0,10,20,30},
    ticklabel style={font=\scriptsize}, label style={font=\scriptsize},
    ylabel={\% of requests}, ylabel shift=-3pt,
    axis lines*=left, axis line style={ohInk}, tick style={ohInk},
    ymajorgrids, grid style={black!10, line width=0.3pt},
    xminorticks=false,
  },
}%
% \ohHist{key}: one method's histogram
\def\ohHist#1{\addplot[const plot, fill=oh#1, fill opacity=0.22, draw=oh#1, line width=0.7pt, forget plot]
  table[x=edge, y=#1] {figures/hist_\ohSet_\ohMetric.dat} \closedcycle;}%
\def\ohSwatch#1{\tikz[baseline=0.1ex]\draw[draw=oh#1, line width=0.7pt, fill=oh#1!40!white] (0,0) rectangle (0.3,0.17);}%
% \ohRow{key}: a row of the table, the method's swatch, p50 and p99
\def\ohRow#1{\ohSwatch{#1} & \ohv{#1}{fifty} & \ohv{#1}{ninetynine} \\}%
\ifdefined\ohTab\else\newsavebox\ohTab\fi
% \ohChart{metric}{x label}{xmax}: the histogram is as tall as the table beside it
\def\ohChart#1#2#3{%
  \def\ohMetric{#1}%
  \sbox\ohTab{\scriptsize\setlength{\tabcolsep}{3pt}\renewcommand{\arraystretch}{1.35}%
    \begin{tabular}{@{}crr@{}}
      \toprule
      & p50 (s) & p99 (s) \\
      \midrule
      \ohRow{nockpt}\ohRow{fat}\ohRow{prmin}%
      \ohSwatch{bounded} & \textbf{\ohv{bounded}{fifty}} & \textbf{\ohv{bounded}{ninetynine}} \\
      \bottomrule
    \end{tabular}}%
  \begin{tikzpicture}
    \begin{axis}[oh panel, height=\ht\ohTab+\dp\ohTab, xmode=log, xmax=#3, xlabel={#2}, name=ax]
      \ohHist{nockpt}\ohHist{fat}\ohHist{prmin}\ohHist{bounded}
    \end{axis}
    \node[anchor=south west, inner sep=0pt] (tab) at ([xshift=0.3cm]ax.south east) {\usebox\ohTab};
    \node[anchor=south, font=\scriptsize, inner sep=0pt]
      at ([yshift=0.32cm]$(ax.north west)!0.5!(tab.east |- ax.north)$)
      {\ohSwatch{nockpt}\,No snapshots\hspace{0.7em}\ohSwatch{fat}\,Fat image\hspace{0.7em}%
       \ohSwatch{prmin}\,Per-request\hspace{0.7em}\ohSwatch{bounded}\,\textbf{\sys (ours)}};
  \end{tikzpicture}}%
\begin{subfigure}[t]{0.49\textwidth}
  \centering
  \ohChart{init}{Initialization time (s)}{100}

  \vspace{-1mm}
  \caption{Initialization time (the shorter, the better)}
  \edef\ohLabel{\noexpand\label{fig:\ohSet_methods_init}}\ohLabel
\end{subfigure}\hfill
\begin{subfigure}[t]{0.49\textwidth}
  \centering
  \ohChart{e2e}{End-to-end time (s)}{1000}

  \vspace{-1mm}
  \caption{End-to-end time (the shorter, the better)}
  \edef\ohLabel{\noexpand\label{fig:\ohSet_methods_e2e}}\ohLabel
\end{subfigure}%
\endgroup

%% file: figures/latency_hist_data.tex
% Generated by experiments/methods/export_tex.py. Do not edit.
\expandafter\def\csname lh-test-n\endcsname{1535}
\expandafter\def\csname lh-test-init-ymax\endcsname{30}
\expandafter\def\csname lh-test-init-xmin\endcsname{0.222}
\expandafter\def\csname lh-test-init-xmax\endcsname{75.89}
\expandafter\def\csname lh-test-init-nockpt-fifty\endcsname{13.9}
\expandafter\def\csname lh-test-init-nockpt-ninety\endcsname{41.8}
\expandafter\def\csname lh-test-init-nockpt-ninetyfive\endcsname{48.7}
\expandafter\def\csname lh-test-init-nockpt-ninetynine\endcsname{57.5}
\expandafter\def\csname lh-test-init-fat-fifty\endcsname{3.75}
\expandafter\def\csname lh-test-init-fat-ninety\endcsname{12.6}
\expandafter\def\csname lh-test-init-fat-ninetyfive\endcsname{15.2}
\expandafter\def\csname lh-test-init-fat-ninetynine\endcsname{19.3}
\expandafter\def\csname lh-test-init-prmin-fifty\endcsname{2.44}
\expandafter\def\csname lh-test-init-prmin-ninety\endcsname{3.71}
\expandafter\def\csname lh-test-init-prmin-ninetyfive\endcsname{4.67}
\expandafter\def\csname lh-test-init-prmin-ninetynine\endcsname{5.88}
\expandafter\def\csname lh-test-init-bounded-fifty\endcsname{0.60}
\expandafter\def\csname lh-test-init-bounded-ninety\endcsname{1.06}
\expandafter\def\csname lh-test-init-bounded-ninetyfive\endcsname{1.57}
\expandafter\def\csname lh-test-init-bounded-ninetynine\endcsname{4.22}
\expandafter\def\csname lh-test-init-preqv-fifty\endcsname{2.38}
\expandafter\def\csname lh-test-init-preqv-ninety\endcsname{3.64}
\expandafter\def\csname lh-test-init-preqv-ninetyfive\endcsname{4.60}
\expandafter\def\csname lh-test-init-preqv-ninetynine\endcsname{5.92}
\expandafter\def\csname lh-test-init-pr-fifty\endcsname{0.53}
\expandafter\def\csname lh-test-init-pr-ninety\endcsname{0.64}
\expandafter\def\csname lh-test-init-pr-ninetyfive\endcsname{0.74}
\expandafter\def\csname lh-test-init-pr-ninetynine\endcsname{0.94}
\expandafter\def\csname lh-test-init-readie-fifty\endcsname{0.45}
\expandafter\def\csname lh-test-init-readie-ninety\endcsname{0.75}
\expandafter\def\csname lh-test-init-readie-ninetyfive\endcsname{0.88}
\expandafter\def\csname lh-test-init-readie-ninetynine\endcsname{1.70}
\expandafter\def\csname lh-test-e2e-ymax\endcsname{30}
\expandafter\def\csname lh-test-e2e-xmin\endcsname{0.233}
\expandafter\def\csname lh-test-e2e-xmax\endcsname{603.49}
\expandafter\def\csname lh-test-e2e-nockpt-fifty\endcsname{15.2}
\expandafter\def\csname lh-test-e2e-nockpt-ninety\endcsname{50.6}
\expandafter\def\csname lh-test-e2e-nockpt-ninetyfive\endcsname{61.7}
\expandafter\def\csname lh-test-e2e-nockpt-ninetynine\endcsname{142.9}
\expandafter\def\csname lh-test-e2e-fat-fifty\endcsname{4.79}
\expandafter\def\csname lh-test-e2e-fat-ninety\endcsname{24.7}
\expandafter\def\csname lh-test-e2e-fat-ninetyfive\endcsname{33.0}
\expandafter\def\csname lh-test-e2e-fat-ninetynine\endcsname{126.9}
\expandafter\def\csname lh-test-e2e-prmin-fifty\endcsname{3.26}
\expandafter\def\csname lh-test-e2e-prmin-ninety\endcsname{15.7}
\expandafter\def\csname lh-test-e2e-prmin-ninetyfive\endcsname{24.8}
\expandafter\def\csname lh-test-e2e-prmin-ninetynine\endcsname{129.6}
\expandafter\def\csname lh-test-e2e-bounded-fifty\endcsname{1.75}
\expandafter\def\csname lh-test-e2e-bounded-ninety\endcsname{9.00}
\expandafter\def\csname lh-test-e2e-bounded-ninetyfive\endcsname{21.4}
\expandafter\def\csname lh-test-e2e-bounded-ninetynine\endcsname{123.0}
\expandafter\def\csname lh-test-e2e-preqv-fifty\endcsname{3.19}
\expandafter\def\csname lh-test-e2e-preqv-ninety\endcsname{15.3}
\expandafter\def\csname lh-test-e2e-preqv-ninetyfive\endcsname{24.2}
\expandafter\def\csname lh-test-e2e-preqv-ninetynine\endcsname{125.0}
\expandafter\def\csname lh-test-e2e-pr-fifty\endcsname{1.56}
\expandafter\def\csname lh-test-e2e-pr-ninety\endcsname{12.8}
\expandafter\def\csname lh-test-e2e-pr-ninetyfive\endcsname{21.6}
\expandafter\def\csname lh-test-e2e-pr-ninetynine\endcsname{125.6}
\expandafter\def\csname lh-test-e2e-readie-fifty\endcsname{1.40}
\expandafter\def\csname lh-test-e2e-readie-ninety\endcsname{7.78}
\expandafter\def\csname lh-test-e2e-readie-ninetyfive\endcsname{16.9}
\expandafter\def\csname lh-test-e2e-readie-ninetynine\endcsname{94.9}
\expandafter\def\csname lh-agent-n\endcsname{832}
\expandafter\def\csname lh-agent-init-ymax\endcsname{25}
\expandafter\def\csname lh-agent-init-xmin\endcsname{0.234}
\expandafter\def\csname lh-agent-init-xmax\endcsname{99.196}
\expandafter\def\csname lh-agent-init-nockpt-fifty\endcsname{14.9}
\expandafter\def\csname lh-agent-init-nockpt-ninety\endcsname{50.3}
\expandafter\def\csname lh-agent-init-nockpt-ninetyfive\endcsname{57.3}
\expandafter\def\csname lh-agent-init-nockpt-ninetynine\endcsname{73.0}
\expandafter\def\csname lh-agent-init-fat-fifty\endcsname{3.48}
\expandafter\def\csname lh-agent-init-fat-ninety\endcsname{13.4}
\expandafter\def\csname lh-agent-init-fat-ninetyfive\endcsname{18.1}
\expandafter\def\csname lh-agent-init-fat-ninetynine\endcsname{41.2}
\expandafter\def\csname lh-agent-init-prmin-fifty\endcsname{2.54}
\expandafter\def\csname lh-agent-init-prmin-ninety\endcsname{5.51}
\expandafter\def\csname lh-agent-init-prmin-ninetyfive\endcsname{9.94}
\expandafter\def\csname lh-agent-init-prmin-ninetynine\endcsname{43.1}
\expandafter\def\csname lh-agent-init-bounded-fifty\endcsname{0.65}
\expandafter\def\csname lh-agent-init-bounded-ninety\endcsname{1.97}
\expandafter\def\csname lh-agent-init-bounded-ninetyfive\endcsname{7.07}
\expandafter\def\csname lh-agent-init-bounded-ninetynine\endcsname{37.1}
\expandafter\def\csname lh-agent-init-preqv-fifty\endcsname{2.25}
\expandafter\def\csname lh-agent-init-preqv-ninety\endcsname{3.59}
\expandafter\def\csname lh-agent-init-preqv-ninetyfive\endcsname{4.37}
\expandafter\def\csname lh-agent-init-preqv-ninetynine\endcsname{5.98}
\expandafter\def\csname lh-agent-init-pr-fifty\endcsname{0.43}
\expandafter\def\csname lh-agent-init-pr-ninety\endcsname{0.63}
\expandafter\def\csname lh-agent-init-pr-ninetyfive\endcsname{0.64}
\expandafter\def\csname lh-agent-init-pr-ninetynine\endcsname{0.85}
\expandafter\def\csname lh-agent-init-readie-fifty\endcsname{0.45}
\expandafter\def\csname lh-agent-init-readie-ninety\endcsname{0.74}
\expandafter\def\csname lh-agent-init-readie-ninetyfive\endcsname{0.79}
\expandafter\def\csname lh-agent-init-readie-ninetynine\endcsname{1.10}
\expandafter\def\csname lh-agent-e2e-ymax\endcsname{25}
\expandafter\def\csname lh-agent-e2e-xmin\endcsname{0.355}
\expandafter\def\csname lh-agent-e2e-xmax\endcsname{544.04}
\expandafter\def\csname lh-agent-e2e-nockpt-fifty\endcsname{16.1}
\expandafter\def\csname lh-agent-e2e-nockpt-ninety\endcsname{61.4}
\expandafter\def\csname lh-agent-e2e-nockpt-ninetyfive\endcsname{77.2}
\expandafter\def\csname lh-agent-e2e-nockpt-ninetynine\endcsname{176.8}
\expandafter\def\csname lh-agent-e2e-fat-fifty\endcsname{4.58}
\expandafter\def\csname lh-agent-e2e-fat-ninety\endcsname{31.6}
\expandafter\def\csname lh-agent-e2e-fat-ninetyfive\endcsname{60.6}
\expandafter\def\csname lh-agent-e2e-fat-ninetynine\endcsname{184.6}
\expandafter\def\csname lh-agent-e2e-prmin-fifty\endcsname{3.28}
\expandafter\def\csname lh-agent-e2e-prmin-ninety\endcsname{16.8}
\expandafter\def\csname lh-agent-e2e-prmin-ninetyfive\endcsname{25.1}
\expandafter\def\csname lh-agent-e2e-prmin-ninetynine\endcsname{54.4}
\expandafter\def\csname lh-agent-e2e-bounded-fifty\endcsname{1.51}
\expandafter\def\csname lh-agent-e2e-bounded-ninety\endcsname{9.19}
\expandafter\def\csname lh-agent-e2e-bounded-ninetyfive\endcsname{16.9}
\expandafter\def\csname lh-agent-e2e-bounded-ninetynine\endcsname{45.5}
\expandafter\def\csname lh-agent-e2e-preqv-fifty\endcsname{2.80}
\expandafter\def\csname lh-agent-e2e-preqv-ninety\endcsname{13.6}
\expandafter\def\csname lh-agent-e2e-preqv-ninetyfive\endcsname{16.5}
\expandafter\def\csname lh-agent-e2e-preqv-ninetynine\endcsname{36.1}
\expandafter\def\csname lh-agent-e2e-pr-fifty\endcsname{1.15}
\expandafter\def\csname lh-agent-e2e-pr-ninety\endcsname{11.6}
\expandafter\def\csname lh-agent-e2e-pr-ninetyfive\endcsname{13.8}
\expandafter\def\csname lh-agent-e2e-pr-ninetynine\endcsname{48.1}
\expandafter\def\csname lh-agent-e2e-readie-fifty\endcsname{1.10}
\expandafter\def\csname lh-agent-e2e-readie-ninety\endcsname{5.75}
\expandafter\def\csname lh-agent-e2e-readie-ninetyfive\endcsname{9.35}
\expandafter\def\csname lh-agent-e2e-readie-ninetynine\endcsname{27.1}

%% file: sections/experiments/init.tex
\subsection{Low Initialization Time}
\label{sec:evaluation_low_init}

This section compares the time taken by each method to initialize a sandbox, at the same storage budget of 15\,GB, on the test set. Per-request at a 15\,GB budget keeps an LRU cache of 8\,GB of snapshots on the host, next to the 7\,GB of packages they load on disk. The other snapshots live in an object store, and a miss fetches one with 20\,ms of latency at 100\,MB/s. \sys plans with a 10\,GB budget, which gives 33 snapshots on the \texttt{python:3.12} image, next to the 5\,GB of packages they load on disk, 15\,GB in all.

We report results in \cref{fig:test_methods} and \cref{table:method_comparison}. \sys spends 0.33\,h initializing the 1,535 requests, which is 22$\times$ less than installing packages per request and importing them (7.31\,h), 6.3$\times$ less than the fat image (2.07\,h) and 3$\times$ less than per-request snapshots (0.98\,h). The fat image avoids installation costs but still imports every package from disk. Per-request snapshots skip the imports, but the 8\,GB cache holds a small share of their 279\,GB, so most requests first fetch their snapshot from the object store. \sys keeps every snapshot on the host, because the 33 shared snapshots fit in the storage budget while all of the 1,308 per-request snapshots do not. \sys also cuts down end-to-end time to 2.96\,h against 9.85\,h, 4.73\,h and 3.88\,h.

\begin{table}[ht]
  \caption{Total initialization and end-to-end time on the test set. The storage budget is set to 15\,GB for all methods.}
  \label{table:method_comparison}
  % \vspace{-2mm}
  
  \centering
  \small
  \begin{sc}
  \input{figures/table_methods_test_bounded.tex}
  \end{sc}
  % \vskip -0.1in
\end{table}

%% file: figures/table_methods_test_bounded.tex
% 1535 test requests that finished ok on every stack; totals in hours.
% Generated by experiments/methods/export_tex.py. Do not edit.
\setlength{\tabcolsep}{3.5pt}%
\begin{tabular}{lrr}
\toprule
Method & Total init (\textnormal{h}) & Total e2e (\textnormal{h}) \\
\midrule
No snapshots & 7.31 & 9.85 \\
Fat image & 2.07 & 4.73 \\
Per-request  & 0.98 & 3.88 \\
\sys (ours) & 0.33 & 2.96 \\
\bottomrule
\end{tabular}

%% file: sections/experiments/snapshot.tex
\subsection{Per-Request Latency at a Fraction of the Storage}
\label{sec:evaluation_low_storage}
This section compares the two methods on the test set without any storage constraints: \sys with 61 snapshots planned at a 20\,GB storage budget against per-request snapshots with every snapshot on the host (279\,GB).

\begin{table}[ht]
  \caption{Total initialization and end-to-end time on the test set without any storage constraints.}
  \label{table:per_request_comparison}
  % \vspace{-2mm}
  \centering
  \small
  \begin{sc}
  \input{figures/table_methods_test_unbounded.tex}
  \end{sc}
  % \vskip -0.1in
\end{table}

We report results in \cref{table:per_request_comparison}. With every snapshot on the host, per-request snapshots give the lowest initialization time any snapshot can, since each request restores exactly the packages it needs. \sys matches it, 0.22\,h against 0.23\,h, at 14$\times$ less storage.

%% file: figures/table_methods_test_unbounded.tex
% 1535 test requests that finished ok on every stack; totals in hours.
% Generated by experiments/methods/export_tex.py. Do not edit.
\setlength{\tabcolsep}{3.5pt}%
\begin{tabular}{lrr}
\toprule
Method & Total init (\textnormal{h}) & Total e2e (\textnormal{h}) \\
\midrule
Per-request & 0.23 & 3.13 \\
\sys (ours) & 0.22 & 2.48 \\
\bottomrule
\end{tabular}

%% file: sections/experiments/unseen.tex
\subsection{Robust to Unseen Workloads}
\label{sec:evaluation_random_starts}

This section compares the same methods as in \cref{sec:evaluation_low_init} on the out-of-distribution set, generated without the knowledge of the requests \sys plans on. \sys serves it with the snapshots it planned using the planning set, and installs or imports any residual packages. We give each unseen request its own snapshot, a hypothetical case (\emph{Per-request} in \cref{sec:evaluation_low_init}), since a platform cannot snapshot code before it sees it.

We report results in \cref{fig:agent_methods} and \cref{table:method_random_comparison}. \sys initializes the 832 requests in 0.43\,h, which is 2$\times$ less than even the per-request snapshots (0.87\,h), 3.3$\times$ less than the fat image and 10.7$\times$ less than having no snapshots or packages installed. Its end-to-end time is 1.07\,h, against 1.69\,h, 3.74\,h and 6.28\,h. \sys's mean initialization grows from 0.77\,s on the test set to 1.86\,s on the out-of-distribution set, and its p95 from 1.57\,s to 7.07\,s on the same sets.

\begin{table}[ht]
  \caption{Total initialization and end-to-end time on the out-of-distribution set.}
  \label{table:method_random_comparison}
  % \vspace{-2mm}
  \centering
  \small
  \begin{sc}
  \input{figures/table_methods_agent_bounded.tex}
  \end{sc}
  % \vskip -0.1in
\end{table}

%% file: figures/table_methods_agent_bounded.tex
% 832 agent requests that finished ok on every stack; totals in hours.
% Generated by experiments/methods/export_tex.py. Do not edit.
\setlength{\tabcolsep}{3.5pt}%
\begin{tabular}{lrr}
\toprule
Method & Total init (\textnormal{h}) & Total e2e (\textnormal{h}) \\
\midrule
No snapshots & 4.59 & 6.28 \\
Fat image & 1.40 & 3.74 \\
Per-request  & 0.87 & 1.69 \\
\sys (ours) & 0.43 & 1.07 \\
\bottomrule
\end{tabular}

%% file: sections/experiments/agents.tex
\subsection{Faster Agent Sandbox Execution}
\label{sec:evaluation_agent}

\begin{figure}[t]
  \centering
  \input{figures/fig_agents_runtime.tex}

  \vspace{-4mm}
  \caption{Total end-to-end time over the tasks of each agent benchmark, in its provided sandbox and in \sys (ours).}
  \label{fig:agents_runtime}
\end{figure}
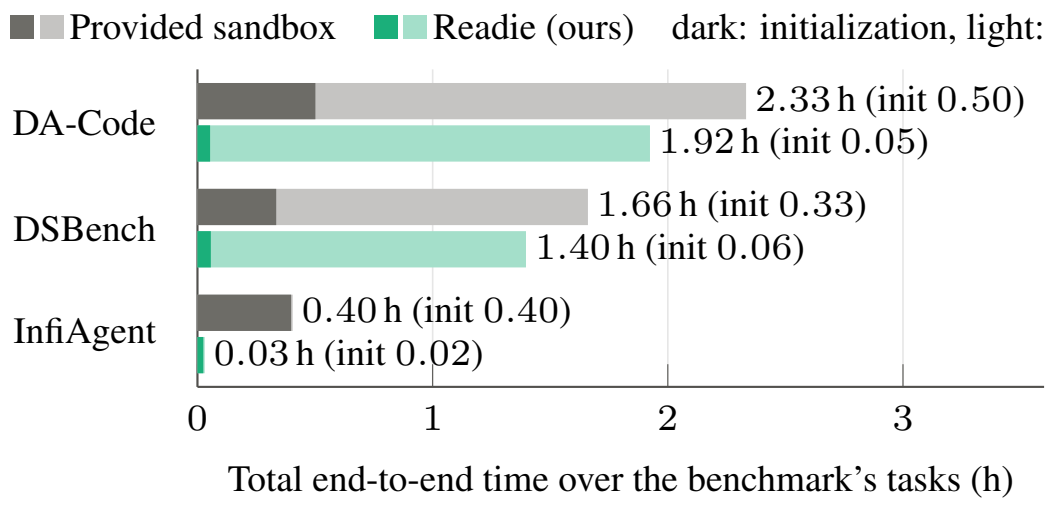

This section measures how much faster an agent can complete the tasks of a benchmark by executing its code on \sys compared to the sandbox environment its benchmark provides. Every turn in which an agent runs code waits for its sandbox, so a faster sandbox shortens each task.

\paragraph{Benchmarks.}
DA-Code~\cite{dacode} tests an agent on data analysis tasks, from data wrangling to machine learning. DSBench~\cite{dsbench} contains data processing and analysis questions and Kaggle-style modeling tasks. InfiAgent~\cite{infiagent} asks closed-form questions about a CSV file, which involve short code snippets. Each benchmark provides a sandbox environment, which we reproduce on a \texttt{python:3.12} image with the packages it lists pre-installed: 16 for DA-Code, 12 for DSBench and 5 for InfiAgent. We record one trajectory per task with Claude Haiku 5.5 as the agent on 419 of DA-Code's, 408 of DSBench's, and 257 of InfiAgent's tasks and then replay its code step by step in both sandboxes. Since the replay runs the same code, the outputs and the pass rate are unaffected.

We report results in \cref{fig:agents_runtime}. Across the three benchmarks, \sys reduces initialization from 1.23\,h to 0.14\,h (9.1$\times$) and end-to-end time from 4.40\,h to 3.35\,h (1.3$\times$). The gain is largest on InfiAgent, 16$\times$ in initialization and 13$\times$ end-to-end, where the initialization time dominates due to shorter tasks. On DA-Code and DSBench, the agent runs longer code, so the initialization times of 0.45\,h and 0.28\,h that \sys saves shorten the end-to-end times by 1.2$\times$.

%% file: figures/fig_agents_runtime.tex
% Total end-to-end time over the tasks of each agent benchmark: one horizontal bar per benchmark and
% sandbox, provided sandbox (grey) above \sys (green); the dark part of each bar is the initialization
% time within it. One column.
% Use: \begin{figure}[t]\centering\input{figures/fig_agents_runtime.tex}\caption{...}\end{figure}; needs \usepackage{pgfplots}.
% Data: figures/agents_runtime.dat, written by experiments/agents/eval/export_tex.py.
\begingroup
\definecolor{arOwn}{HTML}{6D6C67}%
\definecolor{arReadie}{HTML}{1BAF7A}%
\definecolor{arInk}{HTML}{52514E}%
\begin{tikzpicture}
  \begin{axis}[
      width=5.6cm, height=2.1cm, scale only axis,
      xbar, bar width=0.24cm, xmin=0, xmax=3.6,
      ytick={1,2,3}, yticklabels={InfiAgent, DSBench, DA-Code}, ymin=0.5, ymax=3.5,
      y dir=normal,
      ticklabel style={font=\scriptsize}, label style={font=\scriptsize},
      xlabel={Total end-to-end time over the benchmark's tasks (h)},
      axis lines*=left, axis line style={arInk}, tick style={arInk}, ytick style={draw=none},
      xmajorgrids, grid style={black!10, line width=0.3pt},
      clip=false,
      every node near coord/.append style={font=\scriptsize, text=black, anchor=west, inner sep=1.5pt},
      table/y expr=3-\coordindex,
    ]
    % provided sandbox above, Readie below; light bar = end-to-end, dark bar = initialization
    \addplot[fill=arOwn!40, draw=none, bar shift=0.14cm,
      visualization depends on={\thisrow{own_init} \as \arInit},
      nodes near coords={\pgfmathprintnumber[fixed, zerofill, precision=2]{\pgfplotspointmeta}\,h (init \pgfmathprintnumber[fixed, zerofill, precision=2]{\arInit})}]
      table[x=own_e2e] {figures/agents_runtime.dat};
    \addplot[fill=arOwn, draw=none, bar shift=0.14cm] table[x=own_init] {figures/agents_runtime.dat};
    \addplot[fill=arReadie!40, draw=none, bar shift=-0.14cm,
      visualization depends on={\thisrow{readie_init} \as \arInit},
      nodes near coords={\pgfmathprintnumber[fixed, zerofill, precision=2]{\pgfplotspointmeta}\,h (init \pgfmathprintnumber[fixed, zerofill, precision=2]{\arInit})}]
      table[x=readie_e2e] {figures/agents_runtime.dat};
    \addplot[fill=arReadie, draw=none, bar shift=-0.14cm] table[x=readie_init] {figures/agents_runtime.dat};
  \end{axis}
  % key: colour is the sandbox, shade the part of the time
  \node[anchor=south, font=\scriptsize, inner sep=0pt] at ([yshift=4pt, xshift=-4mm]current axis.north) {%
    \textcolor{arOwn}{\rule{1.4ex}{1.4ex}}\,\textcolor{arOwn!40}{\rule{1.4ex}{1.4ex}}\,Provided sandbox\hspace{1em}%
    \textcolor{arReadie}{\rule{1.4ex}{1.4ex}}\,\textcolor{arReadie!40}{\rule{1.4ex}{1.4ex}}\,\sys (ours)\hspace{1em}%
    dark: initialization, light: rest};
\end{tikzpicture}%
\endgroup

%% file: sections/experiments/optimized_ds.tex
\subsection{Optimized for Data Science Workflows}
\label{sec:evaluation_optimized_for_ds}

This section shows how \sys's planned snapshots serve every data science domain with minimal residual import time. We classify the 16 domains in the test set into eight groups, and for each group, visualize the theoretical latency savings and residual import times on \sys against a cold start based on the cost model in \cref{sec:formulation_initialization_time}. All restore and import times are derived from the metadata generated in \cref{sec:implementation_implementation}.

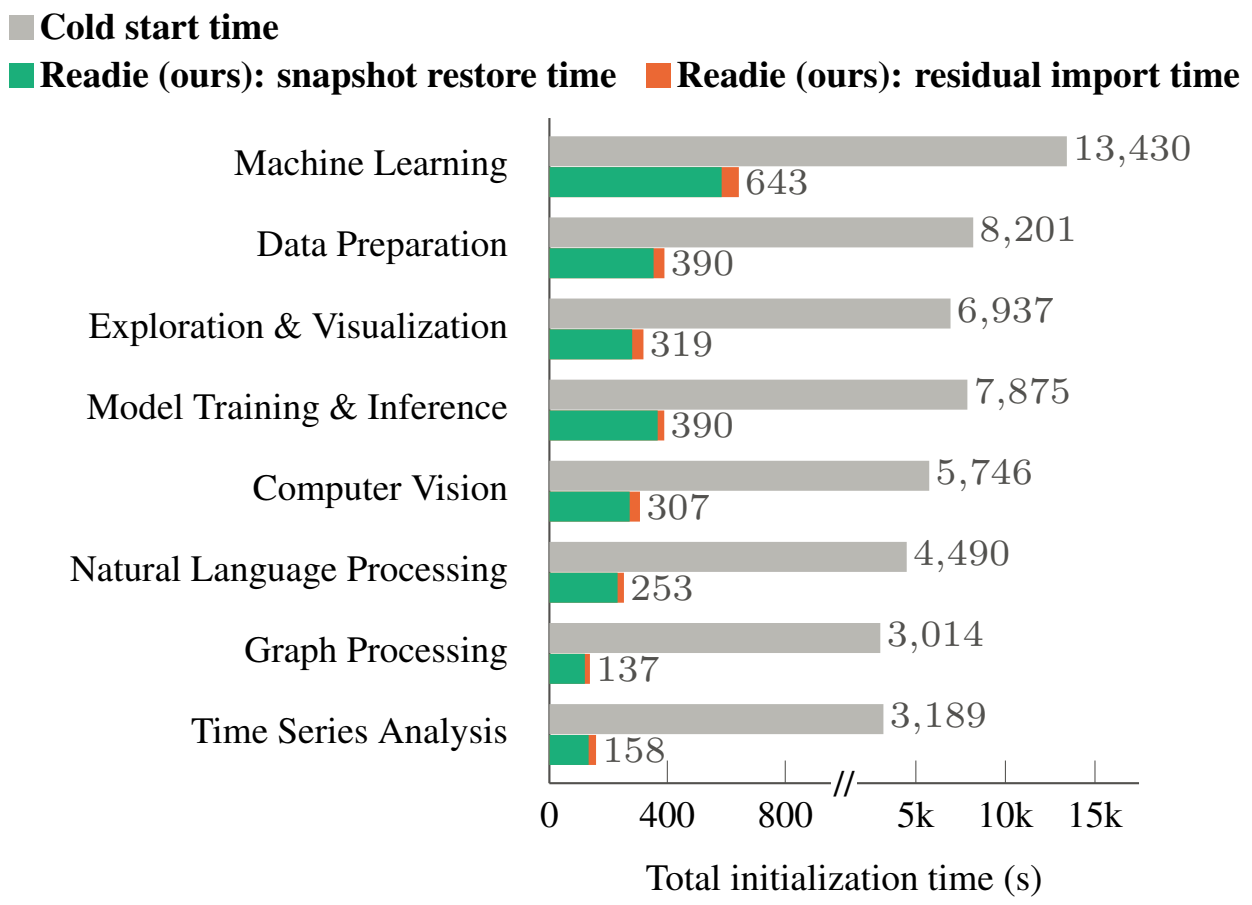
\begin{figure}[t]
  \centering
  \input{figures/fig_domain_residual.tex}

  \vspace{-4mm}
  \caption{Total initialization time of the requests (using the cost model) for each group on the test set: cold start times against the times of \sys (ours), split into snapshot restore/clone time and residual import time. 
  % The time axis is stretched before the break (\texttt{//}) and compressed after it.
  }
  \label{fig:residual_cost}
\end{figure}

We report results in \cref{fig:residual_cost}. \sys reduces the initialization time of requests in each group by 18--22$\times$, and of the test set as a whole from 52,882\,s to 2,596\,s (20$\times$). Restoring the snapshot and importing the residual together take 4.6--5.6\% of cold start time. The residual import time is 255\,s of the 2,596\,s initialization time over the whole test set.

%% file: figures/fig_domain_residual.tex
% Total initialization time per group of domains on the test set: cold start against \sys (checkpoint restore time plus
% residual import time), one column. The time axis is stretched up to the break and compressed after it.
% Use: \begin{figure}[t]\centering\input{figures/fig_domain_residual.tex}\caption{...}\end{figure}; needs \usepackage{pgfplots}.
% Data: figures/domain_residual.dat and domain_residual_data.tex, written by experiments/analysis/export_tex.py.
\begingroup
\input{figures/domain_residual_data.tex}%
\definecolor{drCold}{HTML}{B9B8B3}%
\definecolor{drCost}{HTML}{1BAF7A}%
\definecolor{drResidual}{HTML}{EB6834}%
\definecolor{drInk}{HTML}{52514E}%
\pgfmathtruncatemacro\drTop{\drRows-1}%
\def\drMembers#1{{\tiny\color{drInk}#1}}% the domains a group merges, under its name
\begin{tikzpicture}
  \begin{axis}[
      width=3.9cm, height=4.4cm, scale only axis,
      xbar, bar width=0.37, xmin=0, xmax=1, ymin=-0.6, ymax=\drTop+0.6,
      ytick={0,...,\drTop}, yticklabels/.expand once={\drCoords},
      xtick={\drTicks}, xticklabels/.expand once={\drTickLabels},
      ticklabel style={font=\scriptsize}, yticklabel style={font=\scriptsize, align=right},
      xlabel={Total initialization time (s)}, xlabel style={font=\scriptsize},
      axis lines*=left, axis line style={drInk}, tick style={drInk}, ytick style={draw=none},
      xmajorgrids=false,
      point meta=explicit, nodes near coords, nodes near coords align={horizontal},
      every node near coord/.append style={font=\scriptsize, drInk, inner sep=1pt}, clip=false,
    ]
    \addplot[fill=drCold, draw=none, bar shift=0.19, nodes near coords={\pgfmathprintnumber[fixed, precision=0]{\pgfplotspointmeta}}]
      table[y=y, x=cold, meta=cold_s] {figures/domain_residual.dat};
    \addplot[fill=drResidual, draw=none, bar shift=-0.19, nodes near coords={\pgfmathprintnumber[fixed, precision=0]{\pgfplotspointmeta}}]
      table[y=y, x=readie, meta=readie_s] {figures/domain_residual.dat};
    \addplot[fill=drCost, draw=none, bar shift=-0.19, nodes near coords={}]
      table[y=y, x=cost, meta=readie_s] {figures/domain_residual.dat};
    \node[fill=white, inner sep=0.5pt, font=\scriptsize] at (rel axis cs:\drBreak,0) {//};
  \end{axis}
  % key, centred above the whole figure
  \node[anchor=south, font=\scriptsize\bfseries, inner sep=0pt, align=left] at ([yshift=4pt]current bounding box.north) {%
    \textcolor{drCold}{\rule{1.4ex}{1.4ex}}\,Cold start time\\[1pt]
    \textcolor{drCost}{\rule{1.4ex}{1.4ex}}\,\sys (ours): snapshot restore time\hspace{0.8em}%
    \textcolor{drResidual}{\rule{1.4ex}{1.4ex}}\,\sys (ours): residual import time};
\end{tikzpicture}%
\endgroup

%% file: sections/experiments/algorithm.tex
\subsection{Near-Optimal Snapshot Sets in Less Time}
\label{sec:evaluation_optimal_snapshot}

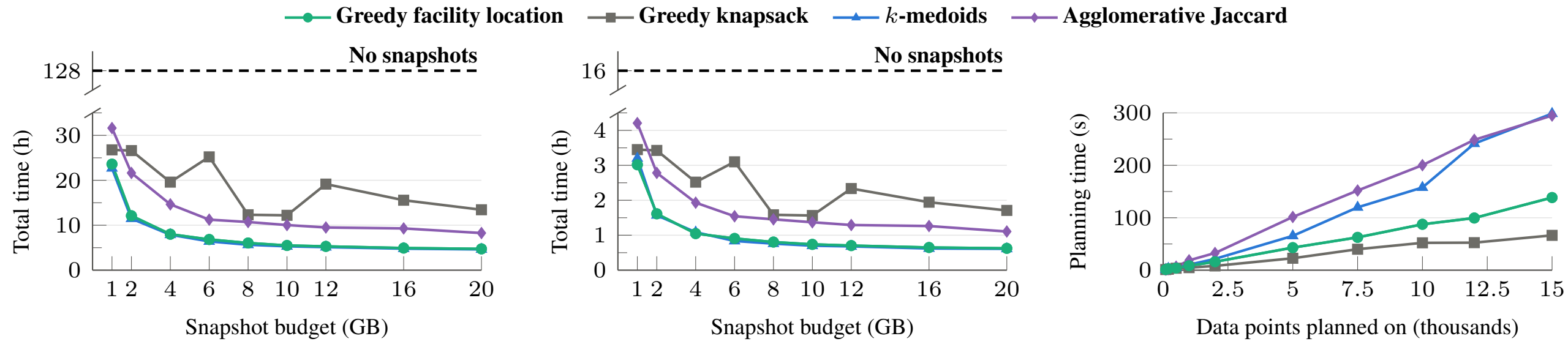
\begin{figure*}[t]
  \centering
  \input{figures/fig_planners_row.tex}
  \caption{Total time to serve the requests of (a) the planning set and (b) the validation set at different storage budgets and (c) the increase in planning time as the number of data points grows.}
  \label{fig:planners}
\end{figure*}

This section compares the greedy facility location algorithm of \sys with other algorithms (\cref{sec:formulation_solving_optimization_problem}) to select snapshot sets, in terms of the overall latency reduction across requests and time taken by the algorithm with increasing requests. For the latency reduction, each algorithm runs on the planning set at budgets from 1\,GB to 20\,GB and generates snapshot sets. The same snapshot sets are then priced on both the planning and validation sets according to the cost model in \cref{sec:formulation_initialization_time}. MILP, which provides the optimal solution, did not resolve within a time limit of 1 hour even for 100 data points when using a CBC solver~\cite{forrest2005cbc} and hence is not considered in this comparison.

We report results for the latency reduction in \cref{fig:planners_a} and \cref{fig:planners_b}. The total cold start time is 128\,h on the planning set and 16.4\,h on the validation set. At a 10\,GB budget, the greedy facility location algorithm brings this to 5.5\,h and 0.74\,h respectively, which is 23$\times$ and 22$\times$ less. The $k$-medoids local search improves on the greedy algorithm by at most 7\% on the planning set, and on the validation set the greedy plan is within 9\% of the $k$-medoids plan at every budget and better at the 1\,GB and 4\,GB budgets. Greedy knapsack spends its budget on closures that overlap the ones it already holds, so its latency does not fall steadily as the budget grows. Agglomerative Jaccard groups sets that share packages, but its unions carry packages that only a few members import, which it pays to restore.

The greedy facility location algorithm is also fast to run (\cref{fig:planners_c}). It plans all 15,299 requests in the planning set in 139\,s, 2.1$\times$ faster than $k$-medoids (299\,s) and agglomerative Jaccard clustering (295\,s). $k$-medoids' improvement of at most 7\% does not justify twice the runtime, and only greedy knapsack is faster (66\,s), but outputs 2.2$\times$ more latency than the greedy facility location.

%% file: figures/fig_planners_row.tex
% Figs (a) planning set, (b) validation set, (c) planning time, side by side across both columns, one shared legend.
% Use: \begin{figure*}\centering\input{figures/fig_planners_row.tex}\caption{...}\end{figure*}; needs \usepackage{pgfplots}.
\begingroup
\def\plWidth{4.2cm}\def\plHeight{1.7cm}\def\plNoLegend{}%
\pgfplotsset{% fewer labelled y ticks, so the short panels stay legible
  pl ticks planner_storage_planning/.style={ytick distance=10, minor y tick num=1},
  pl ticks planner_storage_validation/.style={ytick distance=1, minor y tick num=1},
  pl ticks planner_time/.style={ytick distance=100, minor y tick num=1}}%
\input{figures/fig_planners_legend.tex}\\[2pt]
\begin{subfigure}[t]{0.33\textwidth}\centering
  \input{figures/fig_planners_planning.tex}
  \subcaption{Planning set}\label{fig:planners_a}
\end{subfigure}\hfill
\begin{subfigure}[t]{0.33\textwidth}\centering
  \input{figures/fig_planners_validation.tex}
  \subcaption{Validation set}\label{fig:planners_b}
\end{subfigure}\hfill
\begin{subfigure}[t]{0.33\textwidth}\centering
  \input{figures/fig_planners_time.tex}
  \subcaption{Planning time}\label{fig:planners_c}
\end{subfigure}
\par\vspace{-6pt}% pull the figure caption closer to the sub-captions
\endgroup

%% file: sections/related_work.tex
\section{Related Work}
\label{sec:related}

Our system (\sys) takes a different approach compared to existing techniques for faster code execution.

\paragraph{Per-function snapshots.}
\sys targets unseen code using learned snapshots.
Per-function snapshot techniques instead require the code in advance: they create a snapshot for each function and restore it when the function is invoked~\cite{Prebaking}.
Commercial platforms offer such snapshots~\cite{modal_memory_snapshots,aws_lambda_snapstart}, and later work improves their performance~\cite{Pronghorn,Shin2022Fireworks}.
Since every function needs its own snapshot, these techniques cannot speed up new code, and their footprint grows with the number of functions.

\paragraph{Sharing states across executions.}
% HotSwap~\cite{HotSwap} shares pre-initialized dependencies across functions, avoiding a separate initialized image for each function. Fork-based systems such as AFaaS~\cite{chai2025fork} keep paused processes at useful states and fork them to run code, and Forklift~\cite{yang2024forklift} builds a tree of processes with pre-loaded libraries from a prior distribution, so that it can serve many functions. Like these systems, \sys reuses initialized dependencies across functions. However, it selects a set of snapshots under a storage budget and the snapshot to use for each request, and it keeps its snapshots on disk rather than paying the RAM overhead of forking.
We propose optimized snapshots for container-level isolation, balancing efficiency and security.
Existing systems share live processes in more heuristic ways.
SOCK~\cite{SOCK} and AFaaS~\cite{chai2025fork} fork requests from live pre-initialized processes, and SOCK creates these processes greedily for the current invocation.
Forklift~\cite{yang2024forklift} organizes such processes into a tree, but builds it greedily by usage frequency.
Pagurus~\cite{li2021pagurus} reuses idle containers from other functions, sharing whatever happens to be available.
HotSwap~\cite{HotSwap} keeps dependency images in RAM, built from popular dependency combinations.

\paragraph{Restore time optimizations.}
\sys is orthogonal to other work that speeds up restoration time itself. 
SEUSS~\cite{SEUSS}, Catalyzer~\cite{Catalyzer} and Sabre~\cite{Sabre} accelerate sandbox snapshot restoration, and MITOSIS~\cite{MITOSIS} enables fast distributed forking. Snapshots are often restored lazily, which causes page faults during execution~\cite{REAP}, and REAP~\cite{REAP}, FaaSnap~\cite{FaaSnap} and PASS~\cite{PASS} reduce these page faults.

\paragraph{Other cold start optimizations.}
Container runtimes such as SOCK~\cite{SOCK} and RunD~\cite{Li2022RunD}, and Firecracker~\cite{agache2020firecracker} reduce the overhead of starting and running isolated environments, and could further improve \sys. Pagurus~\cite{li2021pagurus} shares warm containers across functions by repacking them based on function similarity, and FaaSLight~\cite{liu2023faaslight} prunes and lazily loads code that is unlikely to run. Neither restores imported library state from snapshots, as \sys does.

%% file: sections/conclusion.tex
\section{Conclusion}
\label{sec:conclusion}

We propose \sys, a code sandbox system for agents that significantly reduces the startup latency using a small set of learned memory snapshots. 
\sys's contributions include (1) a mathematical model for the sandbox initialization, 
(2) an algorithm that chooses the optimal snapshots under a storage budget, and 
(3) a practical system implementation. 
At the same storage budget, \sys initializes sandboxes 6.3$\times$ faster than a fat image and 3$\times$ faster than per-request snapshots. 
The same snapshots are able to reduce initialization time by 18--22$\times$ across requests in various domains such as data preparation, machine learning, computer vision, and natural language processing.